%% file: main.tex
\documentclass[11pt]{article}

\usepackage[final]{acl}
\usepackage[table]{xcolor}
\usepackage{times}
\usepackage{latexsym}

\usepackage[T1]{fontenc}

\usepackage[utf8]{inputenc}

\usepackage{microtype}

\usepackage{inconsolata}

\usepackage{graphicx}

\usepackage{amsmath}
\usepackage{enumitem}
\usepackage{booktabs}
\usepackage{makecell}
\usepackage{multirow}
\usepackage{tabularx}
\usepackage{longtable}

\usepackage{listings}
\usepackage[most]{tcolorbox}
\usepackage{xcolor}
\lstdefinestyle{python}{
    language=Python,
    basicstyle=\fontsize{7}{8.5}\ttfamily,
    keywordstyle=\color{blue},
    commentstyle=\color{gray},
    stringstyle=\color{black},
    showstringspaces=false,
    breaklines=true,
    breakindent=0pt,
    breakatwhitespace=false,
    escapeinside={(*@}{@*)},
    morecomment=[s]{"""}{"""},
}
\lstdefinestyle{json}{
    basicstyle=\fontsize{7}{8.5}\ttfamily,
    stringstyle=\color{black},
    keywordstyle=\color{blue},
    commentstyle=\color{gray},
    morestring=[b]",
    morekeywords={true,false,null},
    showstringspaces=false,
    breaklines=true,
    breakindent=0pt,
    escapeinside={(*@}{@*)},
}

\definecolor{deepgreen}{RGB}{0, 70, 0}
\definecolor{deepred}{RGB}{182, 32, 22}
\definecolor{jsongreen}{RGB}{0, 153, 0}
\definecolor{lightblue}{RGB}{230, 246, 250}

\newcounter{programbox}
\renewcommand{\theprogrambox}{\arabic{programbox}}

\usepackage{pifont}
\newcommand{\cmark}{\textcolor[rgb]{0.0, 0.6, 0.0}{\ding{51}}} 
\newcommand{\xmark}{\textcolor[rgb]{0.7, 0.0, 0.0}{\ding{55}}} 
\newcommand{\pmark}{\textcolor[rgb]{1,0.647,0}{\ding{51}}}

\title{EnvScaler: Scaling Tool-Interactive Environments for LLM Agent via Programmatic Synthesis}

\author{Xiaoshuai Song, Haofei Chang, Guanting Dong, Yutao Zhu,  Ji-Rong Wen, Zhicheng Dou\thanks{Corresponding Author}\\
  Gaoling School of Artificial Intelligence, Renmin University of China.\\
  \texttt{\{songxiaoshuai,dou\}@ruc.edu.cn}\\
  GitHub: \url{https://github.com/RUC-NLPIR/EnvScaler}
}

\begin{document}
\maketitle
\begin{abstract}
Large language models (LLMs) are expected to be trained to act as agents in various real-world environments, but this process relies on rich and varied tool-interaction sandboxes.
However, access to real systems is often restricted; LLM-simulated environments are prone to hallucinations and inconsistencies; and manually built sandboxes are hard to scale.
In this paper, we propose EnvScaler, an automated framework for scalable tool-interaction environments via programmatic synthesis. 
EnvScaler comprises two components.
First, SkelBuilder constructs diverse environment skeletons through topic mining, logic modeling, and quality evaluation.
Then, ScenGenerator generates multiple task scenarios and rule-based trajectory validation functions for each environment.
With EnvScaler, we synthesize 191 environments and about 7K scenarios, and apply them to Supervised Fine-Tuning (SFT) and Reinforcement Learning (RL) for Qwen3 series models.
Results on three benchmarks show that EnvScaler significantly improves LLMs’ ability to solve tasks in complex environments involving multi-turn, multi-tool interactions.
\end{abstract}

\section{Introduction}
\label{sec:intro}

\begin{figure}[t]
    \centering
    \resizebox{0.48\textwidth}{!}{ \includegraphics{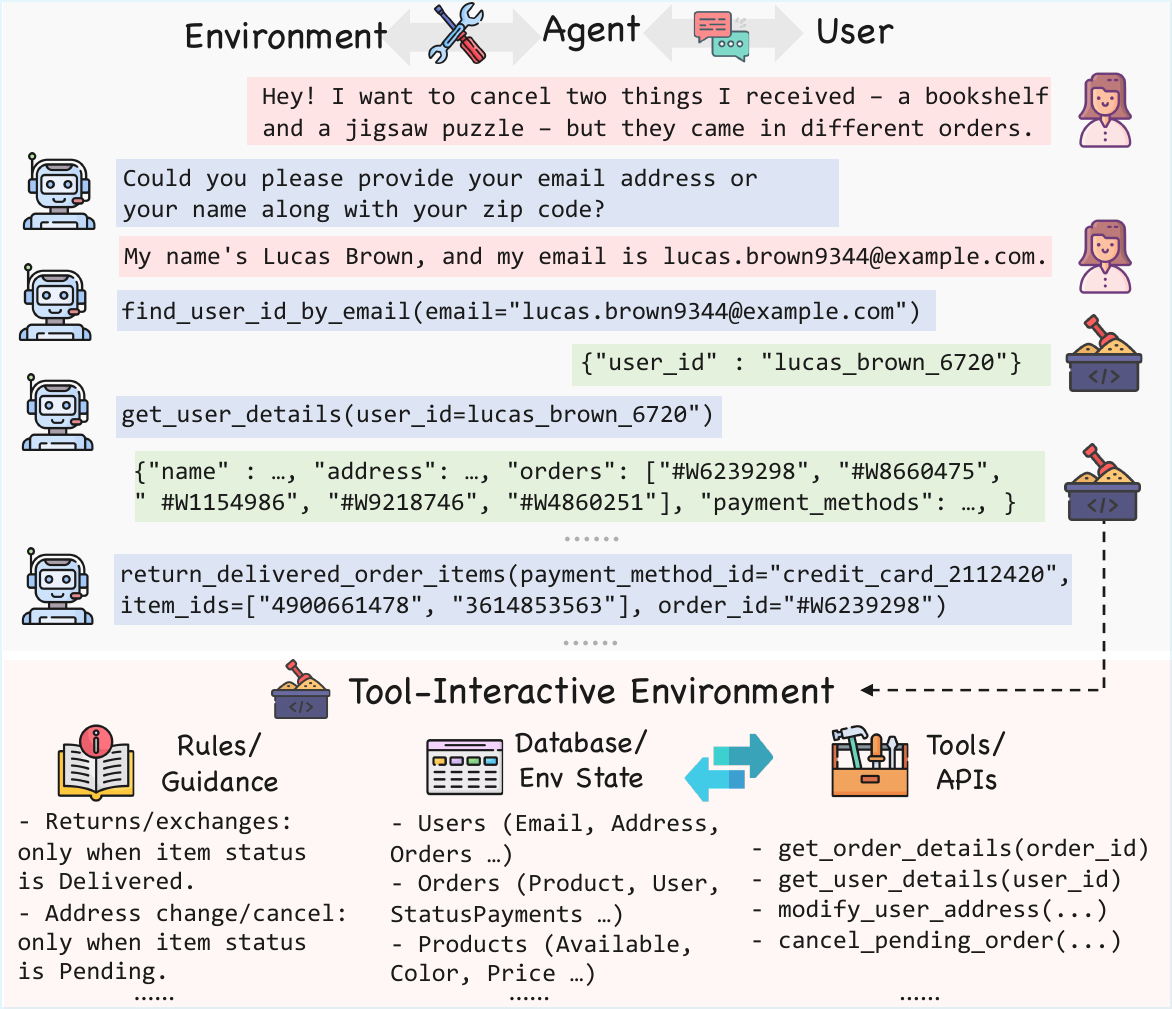}}
    \caption{An illustration of tool-interactive environments. The environment (1) defines rules and provides tool interfaces to the agent; (2) executes the agent’s tool calls to update its state and return results.}
    \label{fig:task_intro}
    \vspace{-0.3cm}
\end{figure}

Large language models (LLMs) are increasingly expected to serve as agents in a wide range of real-world applications, such as modifying orders in e-commerce backends, rescheduling flights via ticketing platforms, or managing documents in a file system~\citep{luo2025large, yao2025taubench, qian2025userbench}.
In these applications, the agent operates within a specific environment (Env), interacting with the user to gather information and invoking tools to query or update the Env’s state, as illustrated in Figure~\ref{fig:task_intro}.
This challenges LLMs to combine dialogue and tool use, adapt actions based on Env feedback, and solve tasks while respecting Env rules over long-horizon trajectories.

To develop such capable LLM agents, scaling up rich and diverse tool-interactive environments is essential. Whether by collecting trajectories followed by imitation learning, or by autonomous exploration and reinforcement learning (RL) within Envs, we hope that exposure to a sufficiently broad range of environments during training will enable LLMs to generalize effectively to unseen environments and scenarios at test time~\citep{huang2025scaling, liu2025deepseek, andrews2025scaling}.
However, as compared in Table~\ref{tab:different_env_type}, real-world environments often have restricted access; LLM-simulated environments also suffer from hallucinations and inconsistencies.
Recently, a series of studies~\citep{patil2025the,yao2025taubench,lu-etal-2025-toolsandbox} build stateful, tool-interactive sandboxes through executable programs, offering advantages in controllability and stability.
Nonetheless, these environments are manually crafted for evaluation purposes, with limited coverage and scalability.
Therefore, a key challenge lies in automating the synthesis and scaling of sandbox environments to support training. It requires creating diverse, high‑quality environments with states, tools, and interaction logic, and designing tasks that align with each environment.

\begin{table}[t]
\small
\centering
\setlength{\tabcolsep}{0.9mm}
\begin{tabular}{lccccc}
\toprule
\textbf{Env Type} & \textbf{\makecell{Scal-\\able}} & \textbf{\makecell{Consis-\\tent}} & \textbf{\makecell{Control-\\lable}} & \textbf{\makecell{Sta-\\ble}} & \textbf{\makecell{Explain-\\able}} \\
\midrule
Real-World & \xmark & \pmark & \xmark & \pmark & \pmark \\
LLM-Simulated & \cmark & \xmark & \pmark & \xmark & \xmark \\
Programmatic & \cmark & \cmark & \cmark & \cmark & \cmark \\
\bottomrule
\end{tabular}
\caption{Key property comparison of three Env types for LLM training.
Scalable: ease of large-scale expansion; Consistent: logical coherence between multiple calls; Controllable: flexibility in modifying Env logic; Stable: reproducible over time; Explainable: transparency of Env logic. 
Symbols denote: \cmark full support, \xmark not supported, \pmark partial or conditional support.}
\label{tab:different_env_type}
\end{table}

\begin{figure}[t]
    \centering
    \resizebox{0.48\textwidth}{!}{
    \includegraphics{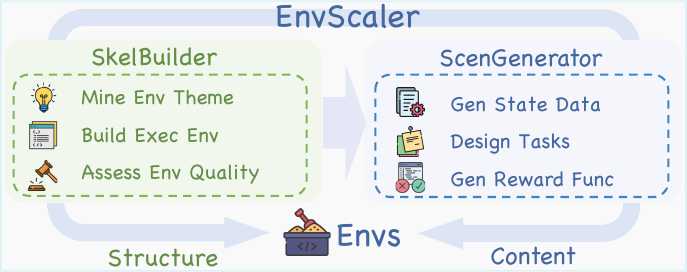}}
    \caption{The overview of EnvScaler.}
    \label{fig:envscaler_overview}
\end{figure}

Several studies have made progress in tackling this challenge, with LLMs used as programmers of environment logic rather than direct simulators.
One approach~\citep{ye2025feedback, sullivan-etal-2025-procedural} focuses solely on tool-layer modeling. It does not model the sandbox’s state, nor consider the interaction logic between tools and the database.
Another approach~\citep{tang2024worldcoder, piriyakulkij2025poe} seeks to programmatically reconstruct environments from existing observations (e.g., trajectories), but inevitably depends on access to pre-existing environments.
Besides, AgentScaler~\citep{fang2025towards} and AutoForge~\citep{cai2025autoforgeautomatedenvironmentsynthesis} rely on pre-collected toolsets or tool documentation, and lack an automated mechanism for assessing environment quality.
Due to these limitations, a notable gap remains in automatically synthesizing and scaling tool-interactive environments without relying on environmental priors or toolsets.
To bridge this gap, we propose \textbf{EnvScaler}, an automated, scalable framework for synthesizing diverse, executable, tool-interactive environments to train LLM agents, as shown in Figure~\ref{fig:envscaler_overview}.

We first introduce \textbf{SkelBuilder} to automate the construction of environmental \textbf{skel}etons, covering topic mining, logic modeling, and assessment. It comprises three modules: 
(1) Task-driven environment discovery: mines diverse environment themes from existing open-source task sets. 
(2) Executable environment construction: starting from an environment description, it plans states and tools, and programmatically implements them into a complete, runnable environment.
(3) Quality inspection: a testing agent sends tool requests, while a checking agent assesses whether executions meet expectations. This process iterates over multiple rounds, with the pass rate indicating environment quality.

To further synthesize multiple task \textbf{scen}arios for each environment, we propose \textbf{ScenGenerator}.
To ensure task relevance and solvability within a given environment and scenario, ScenGenerator first synthesize the environment’s initial database/state, and then derives challenging tasks from the current state.
To achieve rule-based trajectory verification, ScenGenerator generates a set of terminal-state validation functions for each task. 
After the trajectory ends, these functions check whether the final environment state meets the expected conditions, using the functions' pass rate as the reward score.

To validate the effectiveness of EnvScaler, we synthesized 191 environments and about 7K scenarios, applying them to SFT and RL for the Qwen3 series models.
Evaluation on multiple tool-use benchmarks~\citep{patil2025the, yao2025taubench, chen-etal-2025-acebench} shows that EnvScaler significantly enhances LLMs’ ability to solve tasks in complex environments involving multi-turn, multi-tool interactions.
Further analysis of environment coverage, scale, and training strategies provides insights into how synthetic environments promote tool learning and generalization for LLM agents.
\begin{figure*}[t]
    \centering
    \resizebox{0.99\textwidth}{!}{
    \includegraphics{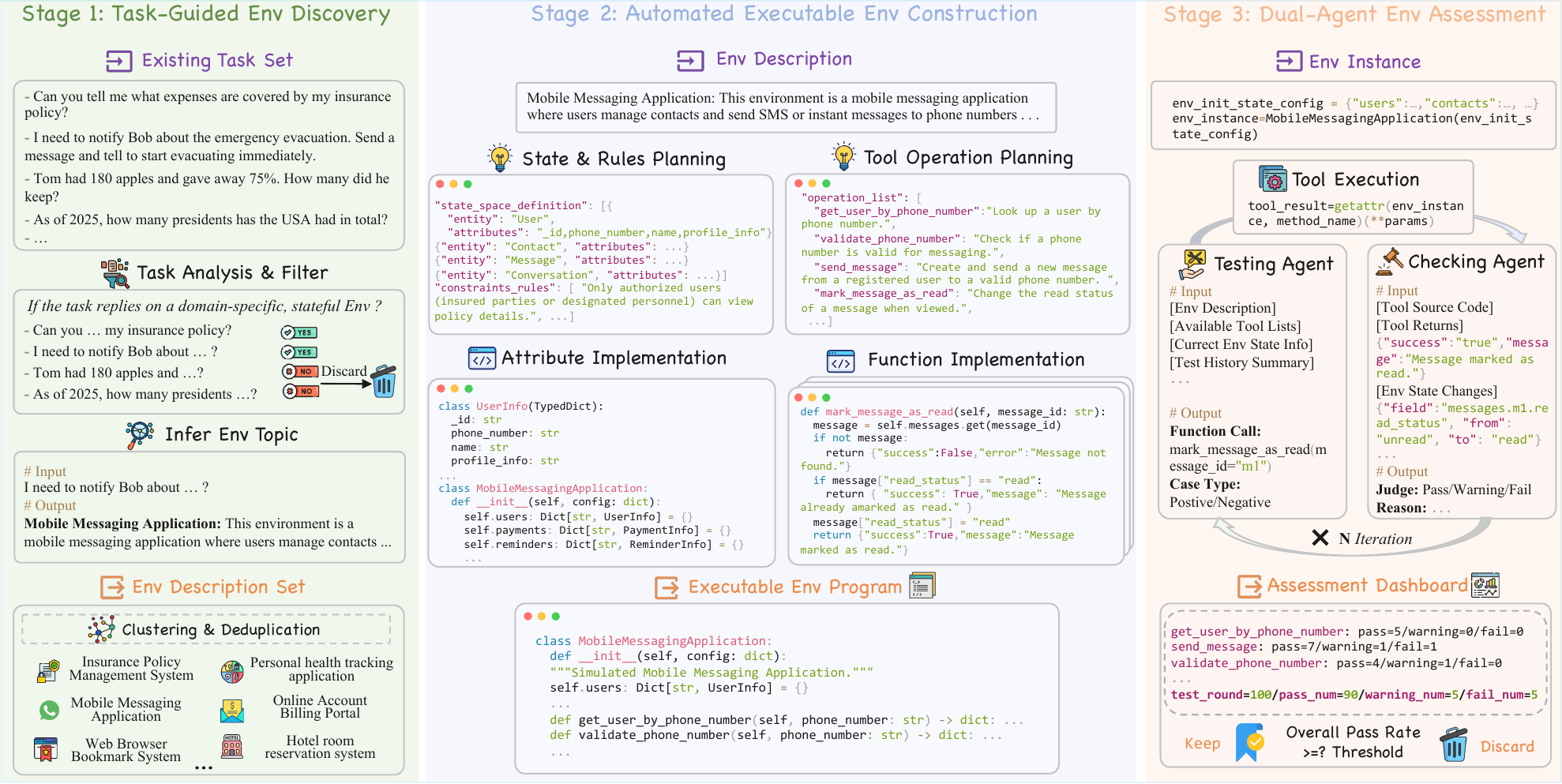}}
    \caption{The overall framework of SkelBuilder.}
    \label{fig:syn_env_main}
\end{figure*}

In summary, we propose EnvScaler for scalable tool-interactive environment synthesis. 
Our contributions are threefold:
(1) We propose SkelBuilder, an automated framework for synthesizing diverse, executable environment skeletons. 
(2) We propose ScenGenerator, a scenario generation pipeline that produces state data, challenging tasks, and rule‑based trajectory verification for each environment. 
(3) Experiments on three benchmarks verify the effectiveness of EnvScaler in improving LLMs' ability to solve tasks in complex environments involving multi-turn, multi-tool interactions.

\section{Related Work}
\label{sec:related_work}

\subsection{Tool Use of LLMs}
\label{subsec:related_work_llm_tool_use}
Many studies aim to improve LLMs' ability to solve tasks with tools~\citep{qu2025tool,luo2025large}.
In this paper, we focus on general tool use across various domain-specific environments~\citep{patil2025the,yao2025taubench,chen-etal-2025-acebench}, rather than tool-integrated reasoning and web information access centered on Python or search tools~\citep{dong2025tool,li2025webthinker}.
Some work have explored the training data and RL strategies from different perspectives~\citep{prabhakar2025apigen,liu2025toolace,xu2025funreasonmttechnicalreportadvanced,zhang2025nemotronresearchtooln1exploringtoolusinglanguage,zhao2025muarlmultiturnuserinteractingagent}.
However, they mainly focus on synthetic static trajectories and cannot support LLMs' self-exploration. For trajectory evaluation, they primarily rely on surface matching, checking whether generated tool names and parameters match references, which is neither sufficient to determine whether the task is truly completed nor able to accommodate multiple equivalent solution paths.
In contrast, we synthesize executable environments and tasks, along with rule-based evaluation grounded in environments’ state, thereby supporting LLMs' training across varied scenarios.

\subsection{Scaling Environments for LLM Agent}
\label{subsec:related_work_scaling_env}
Environments provide agents with action feedback and rewards for interaction and policy optimization.
We focus on tool-interactive environments, where LLM agents can use tools to query environmental information or change the state of the environment.
One line of work~\citep{guo-etal-2024-stabletoolbench, guo2025worldmodellingimproveslanguage, castellani2025synthtoolsframeworkscalingsynthetic, li2025simulatingenvironmentsreasoningmodels} leverages LLMs' reasoning and world knowledge to simulate environments. Although there is no need to build real environments, it is prone to hallucinations and inconsistencies, and lacks transparency and persistent state management.
Another line of work~\citep{tang2024worldcoder, ye2025feedback, fang2025towards, cai2025autoforgeautomatedenvironmentsynthesis} builds sandbox environments through programming. However, they either only model isolated, stateless functions, or rely on environmental priors (e.g., trajectories, toolsets) and lack automatic assessment, which limits scalability and coverage.
Therefore, we propose EnvScaler to enable automatic, scalable environment and scenario synthesis for agent training.

\section{Automated Env Skeleton Synthesis}
\label{sec:syn_env}

\textbf{Overview.}
The goal of SkelBuilder is to construct environments $\{E\}$, where each can be abstracted as a set of three elements: 
$E = \{F_{\text{exec}}, E_{\text{doc}}, \Sigma_{\text{tool}}\}$.
\begin{itemize}
[itemsep=4pt,topsep=0pt,parsep=0pt,leftmargin=9pt]
\item Executable program files $F_{\text{exec}}$: Complete logic implementation of $E$’s states, tools, and rules.
\item Documentation $E_{\text{doc}}$: Provides the agent with introductions or rules about $E$.
\item Tool interface set $\Sigma_{\text{tool}}$: Names, parameters, and descriptions of all tools exposed to the agent, serving as the entry for agent–Env interaction.
\end{itemize}
As shown in Figure~\ref{fig:syn_env_main}, SkelBuilder enables an automated workflow from text resource mining to environment modeling and evaluation.

\subsection{Task-Guided Env Discovery}
\label{subsec:gather_env_topic}
The first step in scaling environments is to collect diverse environment themes.
Unlike manual presetting or derivation from API collections~\citep{fang2025towards}, SkelBuilder mines them from existing text resources.
Considering that studies around SFT have gathered a large and diverse set of tasks that may implicitly contain latent environmental contexts, this inspired us to derive themes through reverse inference from the existing tasks.

Given a task set $T_{\text{exist}} = \{t_1, \dots, t_n\}$, an LLM $M$ first performs binary filtering to retain tasks situated within a domain-specific, stateful environment. For each retained task, $M$ infers the corresponding environment description:
\begin{equation}
\{E'_{\text{des}}\} = \{M(P^{\text{env}}_{\text{infer}}||t) \mid t \in T_\text{exist},M(P^{\text{task}}_{\text{filter}}||t) \}, 
\end{equation}
where $P^{\text{task}}_{\text{filter}}$  and $P^{\text{env}}_{\text{infer}}$ denote prompts for task filtering and environment inference\footnote{We denote the prompt for LLMs by $P$ and use the notation henceforth. The content of $P$ is shown in Appendix \ref{app:env_synth_details}.}. 
The inferred environments are then aggregated and deduplicated by embedding each description and retaining one record from groups of highly similar descriptions, yielding the final diverse, non‑redundant set $ \{E_{\text{des}}\}=\text{Dedup}(\{E'_{\text{des}}\}, \text{sim})$.

\subsection{Automated Executable Env Construction}
\label{subsec:build_env_program}
To transform the environment description into a programmatically modeled environment, we design a three‑stage pipeline. 

\textbf{Logic Planning.}
An LLM enriches the environment description $E_{\text{des}}$, inferring the Env state definition $E_{\text{state}}$, domain rules $E_{\text{rule}}$, and the list of tool operations $ \{E_{\text{tool}_i}\}$. These elements serve as a structured blueprint, with $E_{\text{rule}}$ concatenated with $E_{\text{des}}$ to form the environment documentation $E_{\text{doc}}$: 
\begin{equation}
\begin{aligned}
E_{\text{state}},E_{\text{rule}} &= M(P^{\text{state}}_{\text{plan}}||E_{\text{des}}),\\
\{E_{\text{tool}_i}\} &= M(P^{\text{tool}}_{\text{plan}}|| E_{\text{des}}||E_{\text{state}}||E_{\text{rule}}). 
\end{aligned}
\end{equation}

\textbf{Program Modeling.}
The LLM first converts the planned state space into class attribute definitions $F_{\text{attr}}$. Then, for each tool operation, given the environment rules and class attributes, it generates the corresponding class‑method implementation $F_{\text{meth}_i}$, ensuring consistency with rules and proper state transitions:
\begin{equation}
\begin{aligned}
F_{\text{attr}} &= M(P^{\text{state}}_{\text{exec}}||\ E_{\text{state}}), \\
F_{\text{meth}_{i}} &= M(P^{\text{tool}}_{\text{exec}}||\ E_{\text{rule}}||\ F_{\text{attr}}||\ E_{\text{tool}_i}).
\end{aligned}
\end{equation}

\textbf{Program Assembly.}
The generated code fragments are automatically merged into a complete Python class file $F_{\text{exec}}$, implementing all sandbox logic, where attributes represent environment states and methods represent supported tool operations:
\begin{equation}
F_{\text{exec}} = \text{Merge}(F_{\text{attr}},\{F_{\text{meth}_{i}}\}^m_{i=1}).
\end{equation}
Syntax validity is verified via the abstract syntax tree (AST), with invalid files discarded. AST combined with regex extraction yields all method signatures, forming the tool interface set $\Sigma_{\text{tool}}$\footnote{We provide the example of  $F_{\text{exec}}$, $\Sigma_{\text{tool}}$ in Appendix~\ref{subapp:env_example}.}.

\begin{figure*}[t]
    \centering
    \resizebox{0.99\textwidth}{!}{
    \includegraphics{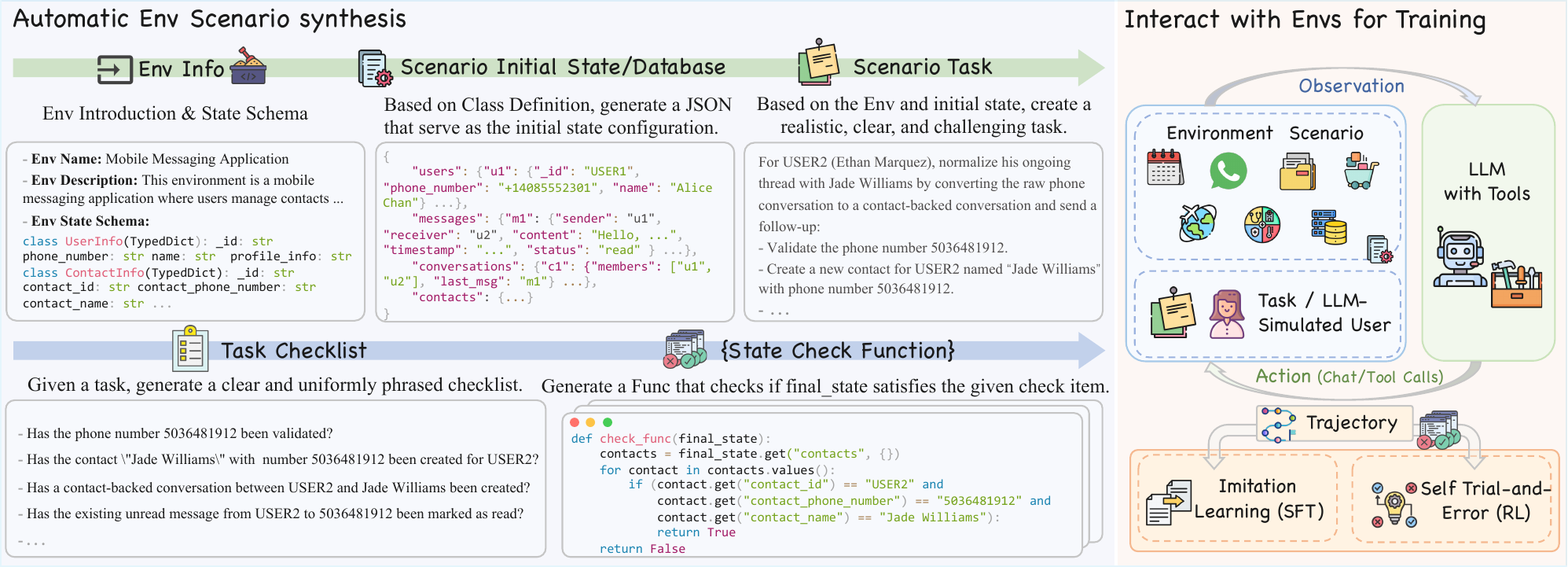}}
    \caption{The overall framework of ScenGenerator.}
    \label{fig:syn_scenario_main}
\end{figure*}

\subsection{Dual-Agent Env Assessment}
\label{subsec:test_env_quality}

Unlike static evaluation via direct LLM scoring, we propose a dual‑agent loop to assess the actual tool-execution performance of the environment.

\textbf{Frontend Testing Agent.}
After instantiating the environment class, the testing LLM agent $M_{\text{test}}$ has no access to its internal implementation. In round $j$, it receives the current environment state $S_j$, and randomly generates a call request, which may be a positive or a negative test case (e.g., invoking a file‑deletion tool to delete a non‑existent file):
\begin{equation}
\text{call}_j = M_{\text{test}}(\Sigma_{\text{tool}}, S_j).
\end{equation}

\textbf{Backend Checking Agent.}
The environment first executes the tool invocation. The checking agent $M_{\text{check}}$, inspects the tool’s source code $F_\text{meth}$, the returned result $R_j$, and state changes before and after execution to judge whether the behavior matches expectations:
\begin{equation}
\begin{aligned}
R_j, S_{j+1} &= \text{Exec}(F_{\text{meth}},\ \text{call}_j,\ S_j),\\
\text{judge}_j &= M_{\text{check}}(\text{call}_j,\ R_j,\ \Delta S_{j \rightarrow j+1}).
\end{aligned}
\end{equation}
The testing and checking agents form a closed loop, iterating for $N$ rounds to cover diverse situations. The average judging pass rate serves as the quantitative metric for the environment’s quality:  
\begin{equation}
\text{score}_{\text{env}} = \frac{1}{N}\sum_{j=1}^{N}\text{judge}_j,
\end{equation}
where environments with $\text{score}_{\text{env}}$ below a predefined threshold are discarded.

\subsection{Practical Analysis}
\label{subsec:env_practice}
In practice, we select API‑Bank~\citep{li-etal-2023-api} and ToolACE~\citep{liu2025toolace} as initial task sources, which have high task retention rates. GPT‑4.1 and Qwen3‑235B‑Instruct‑2507 are used for environment discovery and programming, while GPT‑4.1‑mini and Qwen3‑30B-A3B‑Instruct‑2507 are used for environment assessment. The number of test rounds $N$ is set to 100, with a filtering threshold of 0.85.
In total, we obtain 191 environments. Table~\ref{tab:env_info} and Figure~\ref{fig:env_topic_embedding} presents their statistics, with an average of 18.58 tools and 21.38 state categories per environment, reflecting the complexity of the synthesized environments.

\begin{table}[t]
\small
\centering
\setlength{\tabcolsep}{1.8mm}
\begin{tabular}{lcc}
\toprule
\textbf{Item} & \textbf{Avg.} & \textbf{Med.} \\
\midrule
\textbf{\# Constraint Rules Per Env} & 4.58 & 5 \\
\midrule
\textbf{\# State Category Per Env} & & \\
Level 1 (e.g., user, message, item) & 3.74 & 4 \\
Level 2 per Level 1 (e.g., u\_id, u\_phone) & 5.72 & 5 \\
Total & 21.38 & 21 \\
\midrule
\textbf{\# Tools Per Env} & & \\
Env Information Query (e.g., list\_users) & 10.44 & 10 \\
Env State Change (e.g., send\_message) & 8.14 & 8 \\
Total & 18.58 & 18 \\
\bottomrule
\end{tabular}
\caption{Statistics across 191 synthetic environments.}
\label{tab:env_info}
\end{table}

\section{Automated Env Scenario Synthesis}
\label{sec:syn_scenario}

\textbf{Overview.}
The skeleton of the environment alone is insufficient to support agent interactions. An environment also requires initial state data, tasks, and an evaluation mechanism for assessing agents’ action trajectories. As shown in Figure~\ref{fig:syn_scenario_main}, we propose ScenGenerator to automatically construct multiple task scenarios for each environment.

\subsection{Env Initial State \& Task Generation}
\label{subsec:syn_task}
In stateful environments, tasks are closely tied to state data. For example, an agent cannot cancel an order that does not exist in the environment’s database. 
To ensure that the generated task is both solvable and consistent with the given environment and scenario, ScenGenerator first uses an LLM to generate the environment’s initial state data $S_{\text{init}}$, and then derives challenging tasks based on the environment's initial state:
\begin{equation}
S_{\text{init}} = M(P_{\text{init}}^{\text{gen}}||F_{\text{exec}}||E_{\text{state}}).
\end{equation}

A common task synthesis approach~\citep{fang2025towards,cai2025autoforgeautomatedenvironmentsynthesis} is to walk through tool invocation sequences and infer the corresponding task by reversing the sequence. However, this often yields low-quality tasks, and the existence of multiple valid task solution paths also make the sequence unsuitable as a unique “ground-truth” reference.  
Therefore, we directly prompt the LLM to generate a challenging, scenario-specific task from the initial state $S_0$, tool set $E_{\text{tool}}$, and rules $E_{\text{rule}}$:  
\begin{equation}
\text{task} = M(P_{\text{task}}^{\text{gen}}|| S_{\text{init}}||E_{\text{tool}}||E_{\text{rule}}).
\end{equation}

\subsection{Validation Function Generation}
\label{subsec:syn_check_func}

To achieve rule-based trajectory verification, we first use an LLM to decompose the task into a checklist of verifiable conditions. Then, for each checkpoint, we call the LLM to generate a terminal-state validation function. This function takes the environment’s final state $S_{\text{final}}$ after the trajectory as input and returns ``True'' or ``False'' to indicate whether the condition is satisfied. Finally, the proportion of passed functions is used as the trajectory's reward score. Formally, we have:
\begin{equation}
\begin{aligned}
\{c_k\}_{k=1}^K&=M(P^{\text{check}}_{\text{list}}||\text{task}), \\
f_{c_k}&=M(P^{\text{check}}_{\text{func}}||c_k), \\
\text{reward}&=\frac{1}{K} \sum_{k=1}^{K} \mathbf{1}\left[ f_{c_k}(S_{\text{final}}) = \text{True} \right]. 
\end{aligned}
\end{equation}
Compared with a single boolean judgment, decomposing the task into multiple validation functions not only captures partial completion, but also is easier for the LLM to generate.
Moreover, unlike superficial evaluation that merely checks whether the tool invocation sequence matches a reference sequence, evaluation based on the environment’s final state is agnostic to the solution process and can accommodate multiple valid solution paths.
\begin{table}[t]
    \centering
    \small
    \setlength{\tabcolsep}{0.8mm}
    \begin{tabular}{lccccc}
    \toprule
        ~ & \textbf{\makecell{Qwen3\\-4B(NT)}} & \textbf{\makecell{Qwen3\\-4B}} & \textbf{\makecell{Qwen3\\-8B}} & \textbf{\makecell{Qwen3-8B\\(max@8)}} & \textbf{\makecell{Qwen3\\-30B}} \\
        \midrule
        \multicolumn{6}{l}{\textit{\textbf{Non-Conversation}}} \\
        Score & 37.53 & 53.05 & 57.78 & 69.38 & 67.07 \\
        Avg Step & 21.36 & 12.96 & 13.66 & - & 16.78 \\\midrule
        \multicolumn{6}{l}{\textit{\textbf{Conversation}}} \\
        Score & 46.13 & 55.56 & 58.35 & 74.48 & 63.08 \\ 
        Avg Step & 27.20 & 25.48 & 25.90 & - & 25.38 \\
    \bottomrule
    \end{tabular}
    \caption{Average scores and step counts of LLMs on a random sample of 50 scenarios. Qwen3-30B is Qwen3-30B-A3B-Thinking-2307. NT: Non-Think.}
\label{tab:pre_experiment}
\end{table}

\subsection{Interact with Envs for Training}
\label{subsec:roll_traj}

The agent–environment interaction can be modeled as a Partially Observable Markov Decision Process (POMDP). At time step $t$, the agent $\pi_\theta$ cannot directly access the environment state data $S_t$; instead, it makes decisions based on the observation–action history $H_t$ and the current observation $o_t$:
\begin{equation}
a_t = \pi_\theta(H_t, o_t)\quad o_{t+1},\, S_{t+1} = E(a_t, S_t).
\end{equation}

We consider two interaction settings:\footnote{The trajectory examples are shown in Appendix~\ref{subapp:traj_example}.}

(1) \textbf{Non-Conversation (Non-Conv.)}: The environment directly provides the complete task information to the agent, with the initial observation $o_0 = \{E_\text{doc},\Sigma_\text{tool}, \text{task}\}$. At each step, the agent invokes a tool, and the execution result as the observation, until the agent believes the task is completed or the maximum step limit is reached.

(2) \textbf{Conversation (Conv.)}: The environment additionally includes an LLM-simulated user $\pi_{\text{user}}$. Complete task information is not revealed upfront but must be progressively acquired via agent-user dialogue. Compared with Non-Conv, it expands the agent’s action space to include user interaction. The trajectory ends when the user considers the task completed or the maximum steps are reached.

For SFT, trajectories from the teacher LLM can be directly used as learning targets for the student LLM. For RL, trajectories are converted into rewards using the validation functions, which are then applied for policy optimization.

\subsection{Practical Analysis}
\label{subsec: pre_experiment}

\begin{figure}[t]
    \centering
    \resizebox{0.48\textwidth}{!}{
    \includegraphics{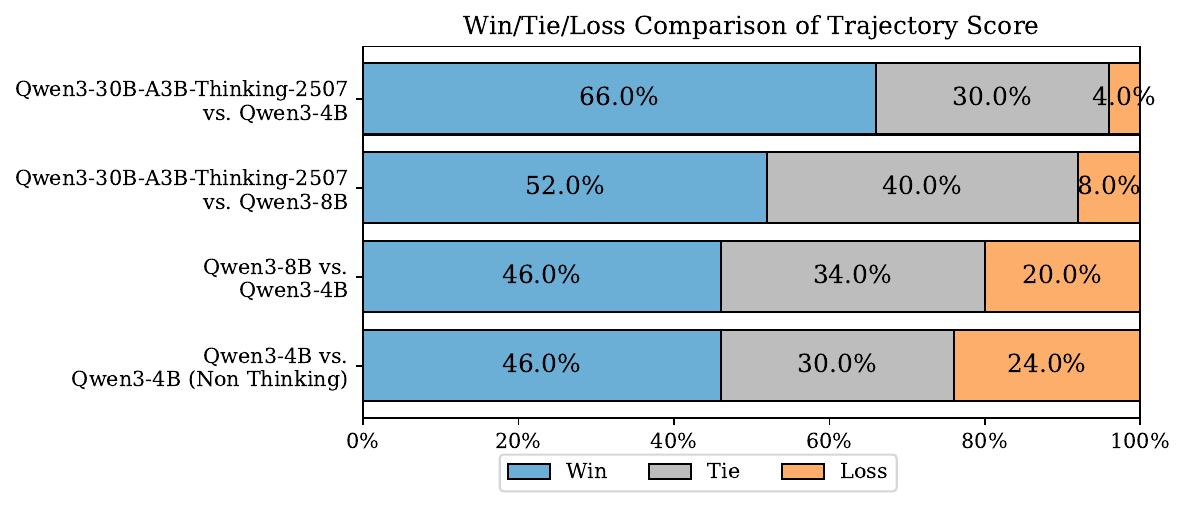}}
    \caption{Pairwise comparison of different LLMs on a random sample of 50 scenarios under Non-Conv setting.}
    \label{fig:win_tie_loss}
\end{figure}

With ScenGenerator, we construct around 7K task scenarios for 191 environments and randomly sample 50 scenarios for a pilot study.
As shown in Table~\ref{tab:pre_experiment}, Qwen3-8B (Thinking) scores 57.78, while Qwen3-4B (Non-Think) scores only 37.53. Besides, the average trajectory is about 15 steps under the Non-Conv setting, increasing to over 25 steps under Conv.
This shows that the synthetic tasks have substantial difficulty and can incentivize LLMs to produce long action trajectories.
Moreover, for the same model, multiple sampling and selecting the highest-scoring trajectory significantly boost the score, suggesting that synthetic tasks offer considerable room for self-exploration and optimization.
Figure~\ref{fig:win_tie_loss} shows that stronger LLMs consistently achieve higher win rates, confirming the validity of the state-check functions in distinguishing and quantifying model performance.

\begin{table*}[t]
\small
\centering
\setlength{\tabcolsep}{0.55mm}
\begin{tabular}{lccccccccccc}
\toprule
\multirow{2}{*}{\textbf{Model}} & \multicolumn{5}{c}{\textbf{BFCL-v3 Multi-Turn}} & \multicolumn{3}{c}{\textbf{Tau-Bench}} & \multicolumn{3}{c}{\textbf{ACEBench-Agent}} \\
\cmidrule(r){2-6}\cmidrule(lr){7-9}\cmidrule(lr){10-12}
 & \textbf{Base}& \textbf{\makecell{Miss-\\Func}} & \textbf{\makecell{Miss-\\Param}} & \textbf{\makecell{Long-\\Context}} & \textbf{Overall} & \textbf{Retail} & \textbf{Airline} & \textbf{Overall} & \textbf{\makecell{Multi-\\Step}} & \textbf{\makecell{Multi-\\Turn}} & \textbf{Overall} \\
\midrule
\multicolumn{12}{l}{\textit{\textbf{Advanced Models}}}\\
GPT-4.1  & 46.00 & 37.50 & 32.00 & 44.50 & 40.00 & 66.95 & 48.0 & 57.48 & 95.00 & 60.00 & 77.50 \\
Qwen3-235B-Thinking-2507  & 60.00 & 35.00 & 34.00 & 54.00 & 45.75 & 67.8 & 46.00 & 56.90 & 85.00 & 63.33 & 74.17 \\
Qwen3-235B-Instruct-2507 & 58.00 & 33.00 & 27.50 & 50.50 & 42.25 & 71.30 & 44.00 & 57.65 & 80.00 & 63.33 & 71.67 \\ 
Kimi-K2-Instruct-0905 & 57.50 & 35.00 & 42.00 & 49.00 & 45.88 & 69.87 & 54.00 & 61.94 & 85.00 & 73.33 & 79.17  \\
\midrule
Qwen3-1.7B  (Thinking)  & 13.50 & 6.00 & 12.50 & 7.00 & \textbf{9.75} & 8.99 & 16.00 & \textbf{12.50} & 35.0 & 28.89 & \textbf{31.95} \\
\rowcolor{lightblue} \quad + EnvScaler (w/ SFT) 
& 24.50 & 11.50 & 20.00 & 16.50 & \textbf{$\text{18.13}_{\textcolor{deepgreen}{\text{+8.38}}}$} 
& 20.87 & 14.00 & \textbf{$\text{17.44}_{\textcolor{deepgreen}{\text{+4.94}}}$ }
& 55.00 & 32.22 & \textbf{$\text{43.61}_{\textcolor{deepgreen}{\text{+11.66}}}$} \\
\rowcolor{lightblue}\quad + EnvScaler (w/ SFT\&RL) 
& 31.50 & 17.00 & 20.50 & 23.00 
& \textbf{$\text{23.00}_{\textcolor{deepgreen}{\text{+13.25}}}$} 
& 18.55 & 14.00 & \textbf{$\text{16.28}_{\textcolor{deepgreen}{\text{+3.78}}}$} 
& 66.67 & 33.33 & \textbf{$\text{50.00}_{\textcolor{deepgreen}{\text{+18.05}}}$} \\
\midrule
Qwen3-4B (Thinking) & 32.00 & 20.00 & 24.00 & 25.50 & \textbf{25.38} & 40.87 & 26.00 & \textbf{33.44} & 58.33 & 52.22 & \textbf{55.28}  \\
\rowcolor{lightblue}\quad + EnvScaler (w/ SFT) 
& 47.00 & 24.00 & 31.50 & 37.00 & \textbf{$\text{34.88}_{\textcolor{deepgreen}{\text{+9.50}}}$} 
& 44.35 & 32.00 & \textbf{$\text{38.20}_{\textcolor{deepgreen}{\text{+4.76}}}$} 
& 73.33 & 60.00 & \textbf{$\text{66.67}_{\textcolor{deepgreen}{\text{+11.39}}}$}  \\
\rowcolor{lightblue}\quad + EnvScaler (w/ SFT\&RL) 
& 51.00 & 34.00 & 28.00 & 39.00 & \textbf{$\text{38.00}_{\textcolor{deepgreen}{\text{+12.62}}}$} 
& 48.12 & 34.00 & \textbf{$\text{41.06}_{\textcolor{deepgreen}{\text{+7.62}}}$} 
& 80.00 & 61.11 & \textbf{$\text{70.55}_{\textcolor{deepgreen}{\text{+15.27}}}$}  \\
\midrule
Qwen3-8B (Thinking) & 32.00 & 33.50 & 22.00 & 28.00 & \textbf{28.88} & 46.38 & 30.00 & \textbf{38.19} & 63.33 & 56.67 & \textbf{60.00}  \\
\rowcolor{lightblue}\quad + EnvScaler (w/ SFT) 
& 47.00 & 33.00 & 29.50 & 38.50 & \textbf{$\text{37.00}_{\textcolor{deepgreen}{\text{+8.12}}}$} 
& 48.70 & 34.00 & \textbf{$\text{41.35}_{\textcolor{deepgreen}{\text{+3.16}}}$} 
& 83.33 & 60.00 & \textbf{$\text{71.67}_{\textcolor{deepgreen}{\text{+11.67}}}$}  \\
\rowcolor{lightblue}\quad + EnvScaler (w/ SFT\&RL) 
& 55.50 & 36.00 & 35.00 & 41.00 & \textbf{$\text{41.88}_{\textcolor{deepgreen}{\text{+13.00}}}$} 
& 53.62 & 36.00 & \textbf{$\text{44.81}_{\textcolor{deepgreen}{\text{+6.62}}}$} 
& 85.00 & 60.00 & \textbf{$\text{72.50}_{\textcolor{deepgreen}{\text{+12.50}}}$}  \\
\bottomrule
\end{tabular}
\caption{Performance comparison of models on three benchmarks. We bold the Overall results for each benchmark.}
\label{tab:main_results}
\end{table*}

\section{Experiments}
\label{sec:experiment}

\subsection{Experiment Setup}
\label{subsec:exp_setup}

\textbf{Training.}
We conduct SFT and RL on Qwen3 series models (Thinking Mode)~\citep{yang2025qwen3}.
A total of 140 environments are used for SFT, and the remaining 51 environments are used for RL.
For SFT, we employ Qwen3-30B-A3B-Thinking-2507 as the teacher model, interacting with the environments under two interaction settings, yielding about 9K trajectories in total.
For RL, we apply the Reinforce++ algorithm~\citep{hu2025reinforcestabilizingcriticfreepolicy} under Non-Conv Setting. Detailed implementation are provided in Appendix \ref{subapp:implement_details}.\\
\textbf{Evaluation.}
We use three widely used multi-turn tool-use benchmarks: BFCL-v3 Multi-Turn (abbreviated as BFCL‑MT)~\citep{patil2025the}, Tau-Bench~\citep{yao2025taubench}, and ACEBench-Agent~\citep{chen-etal-2025-acebench}. Each benchmark consists of several domain-specific environments equipped with tools, requiring the LLM to interact with users and invoke tools to solve tasks. Notably, ACEBench uses the \texttt{[func\_name(param)]} prompt format by default; we modify the official code to support LLMs’ native function-calling interface to ensure consistency. 
Detailed introductions are provided in Appendix \ref{subapp:dataset_info} and Table~\ref{tab:eval_dataset_info}.

\subsection{Main Results}
\label{subsec:main_result}

\begin{figure}[t]
    \centering
    \resizebox{0.48\textwidth}{!}{
    \includegraphics{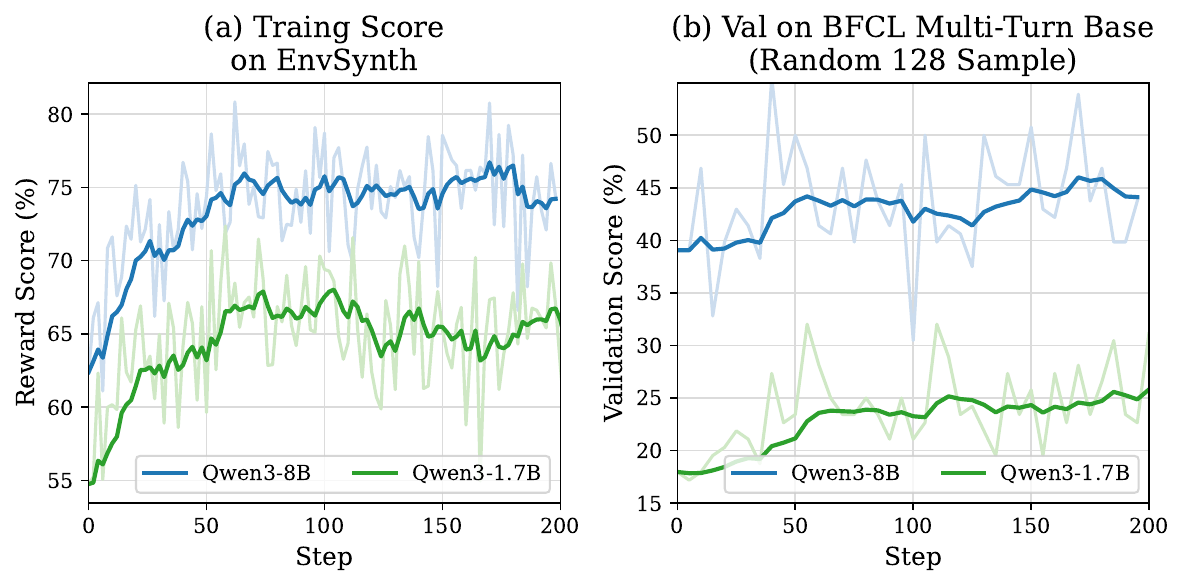}}
    \caption{The RL training and validation curve of Qwen3 in synthetic environments after SFT.}
    \label{fig:sft_rl_curve}
\end{figure}

As shown in Table~\ref{tab:main_results}, using EnvScaler for SFT leads to significant improvements across all benchmarks, and incorporating RL further boosts performance~\footnote{The experiments in direct RL and non-thinking mode are presented in Section~\ref{subapp:direct_rl_result} and Appendix~\ref{subapp: non_think_result}, respectively.}. 
Next, we analyze the results from three aspects:

(1) \textbf{Training Strategy.}
SFT significantly boosts LLMs' performance. Averaged across three models, BFCL-MT improves by 8.67 points, Tau-Bench by 4.29 points, and ACEBench-Agent by 11.57 points.
This can be attributed to the diverse, multi-tool environments and complex tasks synthesized by EnvScaler, which help LLMs strengthen domain adaptability, multi-turn interaction, and multi-tool usage in supervised training. 
Incorporating RL yields further improvements (the RL curve is shown in Figure~\ref{fig:sft_rl_curve}). For example, Qwen3-8B achieves 4.88 and 3.46 point gains on BFCL-MT and Tau-Bench, respectively. This demonstrates that leveraging synthetic environments and state-checking reward signals can guide LLMs to autonomously learn more effective strategies.

(2) \textbf{Model Size.}
SFT delivers stable improvements across models, whereas RL performance is more model‑dependent.
Qwen3‑8B achieves gains across all datasets, whereas Qwen3‑1.7B shows notable improvements on BFCL-MT and ACEBench‑Agent but a slight drop on Tau‑Bench.
The main reason is that large‑scale models possess stronger exploration capabilities during RL, enabling them to extract effective strategies. In contrast, small‑scale models, with weaker foundational abilities, are more susceptible to noisy reward signals and produce lower-quality strategies that struggle to generalize to unseen environments. 

(3) \textbf{Different Benchmarks.}
Comparatively, EnvScaler shows more substantial gains on BFCL-MT and ACEBench-Agent, while improvements on Tau-Bench are relatively limited. Qwen3-1.7B even slightly declines on the most challenging Tau-Bench Airline task.
This is mainly because BFCL-MT and ACEBench-Agent underscore the evaluation of LLM’s ability in multi-turn, multi-tool  collaboration and domain adaptability, while EnvScaler can substantially enhance these capabilities.
In contrast, Tau-Bench focuses on deep reasoning under complex environment’s rules, making such tasks harder for LLMs to learn and generalize.



\begin{table}[t]
\footnotesize
\centering
\setlength{\tabcolsep}{0.7mm}
\begin{tabular}{lccccc}
\toprule
 & \textbf{Base} & \textbf{\makecell{Miss-\\Func}} & \textbf{\makecell{Miss-\\Parm}} & \textbf{\makecell{Long-\\Cont.}} & \textbf{\makecell{Over-\\all}} \\
\midrule
Qwen3-4B & 32.00 & 20.00 & 24.00 & 25.50 & 25.38 \\
\quad+ SFT (Full) & 47.00 & 24.00 & 31.50 & 37.00 & 34.88 \\
\quad+ SFT (top 50\%) & 40.50 & 28.00 & 27.50 & 32.00 & 32.00 \\
\quad+ SFT (bottom 50\%) & 43.00 & 27.50 & 27.50 & 32.00 & 32.50 \\
\quad+ SFT (random 50\%)\,& 42.00 & 26.00 & 26.00 & 31.00 & 31.25 \\
\bottomrule
\end{tabular}
\caption{Qwen3-4B's Performance on BFCL-MT with training on different environment-similarity subsets.}
\label{tab:bfcl_decup}
\end{table}

\begin{figure}[t]
    \centering
    \resizebox{0.48\textwidth}{!}{
    \includegraphics{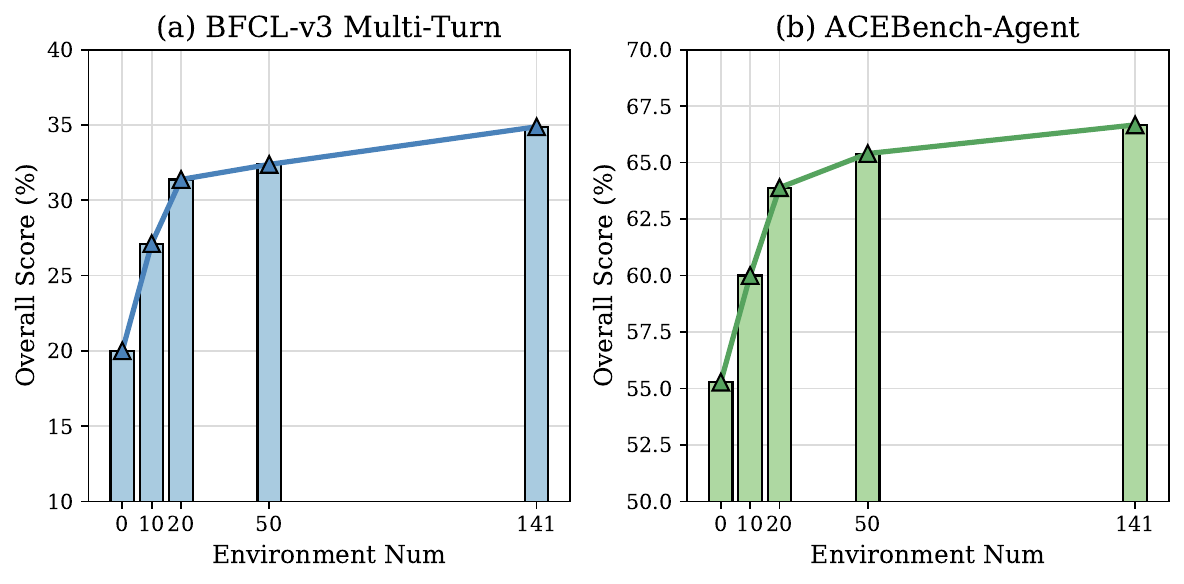}}
    \caption{The change of Qwen3-4B's performance with the scaling of the number of environments for SFT.}
    \label{fig:env_num_sft_effect}
\end{figure}

\subsection{Train-Test Env Similarity Analysis}
To investigate how the similarity between synthetic and test environments affects test performance, we compute the similarity between each synthetic environment and those in BFCL-MT, based on textual embeddings of the environment topic and toolset descriptions.
According to the similarity scores, environments are divided into three groups: (1) the 50\% most similar; (2) the 50\% least similar; and (3) a random 50\% selection.
As shown in Table~\ref{tab:bfcl_decup}, training on either the most similar or least similar subsets yields only minor differences in overall metrics, and both significantly outperform the baseline.
This suggests that the performance gains do not primarily depend on direct similarity between training and test environments, but on the problem‑solving patterns and transferable tool‑use skills learned from different environments.

\begin{table}[t]
\small
\centering
\setlength{\tabcolsep}{1mm}
\begin{tabular}{lccccc}
\toprule
 & \textbf{Base} & \textbf{\makecell{Miss\\Func}} & \textbf{\makecell{Miss\\Parm}} & \textbf{\makecell{Long\\Cont.}} & \textbf{\makecell{Over-\\all}} \\
\midrule
Qwen3-8B & 32.00 & 33.50 & 22.00 & 28.00 & 28.88 \\
\quad + SFT (Non-Conv) & 49.50 & 32.50 & 23.50 & 37.50 & 35.75 \\
\quad + SFT (Conv) & 45.50 & 32.00 & 30.50 & 34.00 & 35.50 \\
\quad + SFT (Full) & 47.00 & 33.00 & 29.50 & 38.50 & 37.00 \\
\bottomrule
\end{tabular}
\caption{Qwen3-8B's Performance on BFCL-MT with
training on different subsets of interaction patterns.}
\label{tab:ablation}
\end{table}

\subsection{Effect of Scaling Environment}
To analyze the effect of scaling the number of training environments on model performance, we sample subsets with varying environment counts from the full SFT dataset and conduct SFT on each subset.
As shown in Figure~\ref{fig:env_num_sft_effect}, Qwen3‑4B’s scores on both the BFCL‑MT and ACEBench‑Agent exhibit a steady upward trend as the number of training environments increases.
The most pronounced improvement occurs when scaling from 0 to 20 environments; although the rate of gain slows beyond this point, performance continues to rise overall.
These results indicate that scaling up the number and diversity of training environments effectively enhances the model’s adaptability across contexts and improves its task‑solving performance.

\begin{table}[t]
\small
\centering
\setlength{\tabcolsep}{1mm}
\begin{tabular}{lccc}
\toprule
   \textbf{\makecell{Model\\(Thinking)}} & \textbf{\makecell{BFCL\\Multi-Turn}} & \textbf{\makecell{ACEBench\\-Agent}} & \textbf{\makecell{Tau-\\Bench}} \\
\midrule
Qwen3-1.7B 
  & 9.75 
  & 12.50 
  & 31.95 \\ 
\quad+ EnvScaler (RL) 
  & $\text{12.12}_{\textcolor{deepgreen}{\text{+2.37}}}$ 
  & $\text{13.08}_{\textcolor{deepgreen}{\text{+0.58}}}$ 
  & $\text{32.22}_{\textcolor{deepgreen}{\text{+0.27}}}$ \\
\midrule
Qwen3-4B
  & 25.38 
  & 33.44 
  & 55.28 \\ 
\quad+ EnvScaler (RL) 
  & $\text{29.38}_{\textcolor{deepgreen}{\text{+4.00}}}$ 
  & $\text{35.15}_{\textcolor{deepgreen}{\text{+1.71}}}$ 
  & $\text{61.25}_{\textcolor{deepgreen}{\text{+5.97}}}$ \\
\midrule
Qwen3-8B 
  & 28.88 
  & 38.19 
  & 60.00 \\ 
\quad+ EnvScaler (RL) 
  & $\text{37.62}_{\textcolor{deepgreen}{\text{+8.74}}}$ 
  & $\text{42.30}_{\textcolor{deepgreen}{\text{+4.11}}}$ 
  & $\text{63.05}_{\textcolor{deepgreen}{\text{+3.05}}}$ \\
\bottomrule
\end{tabular}
\caption{Performance of models trained with direct RL.}
\label{tab:direct_rl}
\end{table}

\begin{figure}[t]
    \centering
    \resizebox{0.48\textwidth}{!}{
    \includegraphics{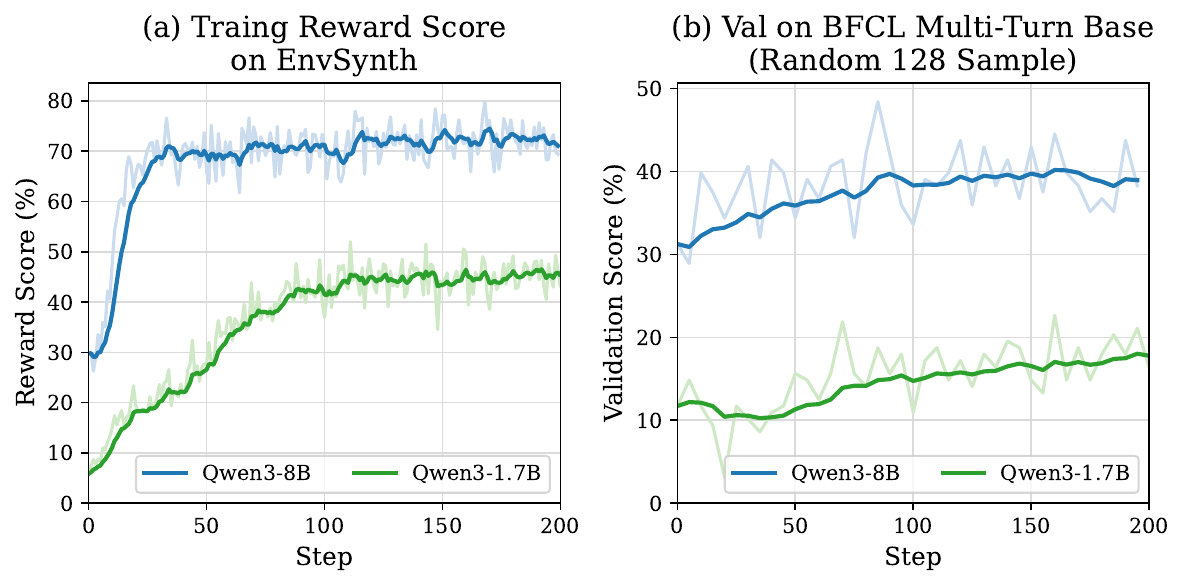}}
    \caption{The Direct RL training and validation curve of Qwen3 in synthetic environments without SFT.}
    \label{fig:direct_rl_curve}
\end{figure}

\subsection{Effect of Interaction Patterns}
To investigate the impact of interaction patterns on model training, we conduct SFT under three data settings: Non‑Conv, Conv, and their combination (Full).
Table~\ref{tab:ablation} reports the BFCL‑MT results for Qwen3‑8B. The Base and Long-Context subsets have complete task information, whereas the Miss‑Func and Miss‑Parm subsets lack necessary tools or parameters, requiring the agent to interact with the user to obtain them.
It can be observed that SFT (Non-Conv) performs better on Base and Long-Context, while SFT (Conv) achieves higher scores on Miss‑Parm. Performance on Miss‑Func does not improve, primarily because the training data lacks samples with missing tool types. 
The SFT (Full) setting yields the highest overall score, highlighting the need to learn both interaction patterns for adaptability in scenarios with either sufficient or incomplete information.

\subsection{Results of Training with Direct RL}
\label{subapp:direct_rl_result}

Table~\ref{tab:direct_rl} shows the performance of models trained with RL directly on synthetic environments without SFT cold start. Figure~\ref{fig:direct_rl_curve} presents the curves during the RL process.
Overall, all models achieve varying degrees of improvement on three benchmarks, indicating that even without SFT initialization, direct RL can still drive self-optimization.
However,  the gains from direct RL strongly depend on model size and exploration capability. Qwen3‑1.7B shows only minor improvement, while Qwen3‑8B achieves the most notable gains. This suggests that larger models can leverage the environment more effectively for exploration and learning high‑quality policies, whereas smaller models, limited in base ability and exploration, are more vulnerable to sparse or noisy rewards.
Besides, compared with ``SFT + RL'' in Table~\ref{tab:main_results}, the overall gains from direct RL remain limited, highlighting that SFT initialization is still important for better policy quality. Combining SFT with RL can more fully exploit the training value of synthetic environments.

\subsection{Quality Analysis of Environment}

\begin{figure}[t]
    \centering
    \resizebox{0.48\textwidth}{!}{
    \includegraphics{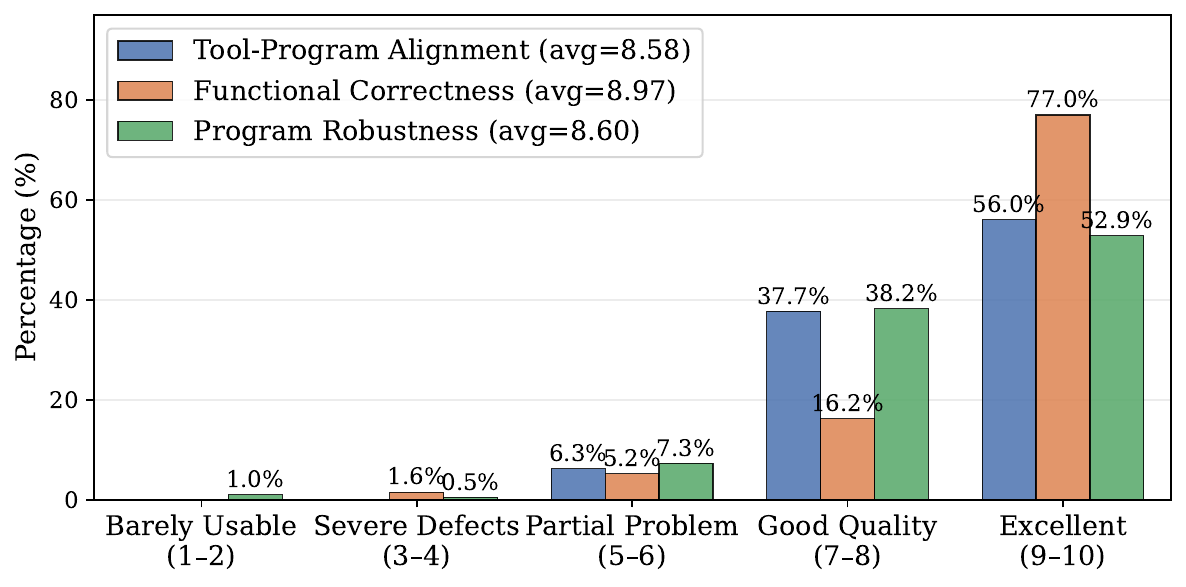}}
    \caption{The score distribution of environment quality.}
    \label{fig:env_quality_score}
\end{figure}

To further evaluate the quality of environments, we decompose the quality evaluation into three dimensions and find that Claude-4.5-Sonnet’s scores align highly with manual judgments:
(1) Tool-Program Alignment: it evaluates whether the semantic definition of the tool is consistent with its program implementation.
(2) Functional Correctness: it evaluates whether the tool behaviors in the environment meet expectations.
(3) Code Robustness: it evaluates whether the environment program handles abnormal inputs and boundary conditions reasonably.
As shown in Figure~\ref{fig:env_quality_score}, the 191 synthesized environments achieve high average scores of 8.5+ across all three dimensions. This proves the overall reliability of the EnvScaler evaluation mechanism.

\subsection{Token Usage and Cost Analysis}
\label{subsec:cost}

\begin{table}[t]
\centering
\small
\setlength{\tabcolsep}{1.6mm}
\begin{tabular}{lcccc}
\toprule
\multirow{2}{*}{\textbf{Stage}} & \multicolumn{3}{c}{\textbf{Tokens}} & \multirow{2}{*}{\textbf{\makecell{Cost\\(\$)}}} \\
\cmidrule(lr){2-4}
 & \textbf{Input} & \textbf{Output} & \textbf{Total} & \\
\midrule
\multicolumn{5}{l}{{\textbf{\textit{Per Environment (SkelBuilder)}}}} \\
\makecell[l]{\textit{Discovery} \\ (GPT-4.1)}       & 1529    & 569    & 2098    & 0.0076 \\
\makecell[l]{\textit{Construction} \\ (GPT-4.1)}      & 57639   & 16655  & 74294   & 0.2485 \\
\makecell[l]{\textit{Accessment} \\ (GPT-4.1-mini)}    & 1709159 & 52706  & 1761865 & 0.7679 \\
Total                                     & 1768327 & 69930  & 1838257 & 1.0240 \\
\midrule
\multicolumn{5}{l}{\textbf{\textit{Per Scenario (ScenGenerator)}}} \\
\makecell[l]{\textit{Gen Database} \\ (GPT-4.1)}       & 5658   & 2205  & 7863   & 0.0289 \\
\makecell[l]{\textit{Gen Task} \\ (GPT-4.1)}           & 2883   & 555   & 3438   & 0.0102 \\
\makecell[l]{\textit{Check Func} \\ (GPT-4.1-mini)}     & 26879  & 8571  & 35450  & 0.0244 \\
Total                                     & 35420  & 11331 & 46751  & 0.0635 \\
\bottomrule
\end{tabular}
\caption{Average token usage and cost of EnvScaler.}
\label{tab:cost_analysis}
\end{table}

We evaluate the average token usage and cost of synthesizing a single environment and a single scenario on a small batch.
The results are shown in Table~\ref{tab:cost_analysis}.
Under the GPT-4.1/mini combination, the average cost per environment is about \$1 and the cost per scenario is about \$0.06. We consider this acceptable. During environment construction, the 100-round dual-agent loop causes high token consumption. If the number of loop rounds is reduced, the cost can be further lowered.

\section{Conclusion}
\label{sec:conclusion}
In this paper, we propose EnvScaler, a scalable framework for synthesizing diverse, executable tool-interactive environments to train LLM agents. 
EnvScaler first constructs environment skeletons through topic mining, programmatic modeling, and dual-agent assessment (SkelBuilder), then synthesizes scenarios by generating initial states, challenging tasks, and rule-based trajectory verification (ScenGenerator). 
Experiments on three benchmarks demonstrate that EnvScaler significantly enhances LLMs’ ability to solve tasks through multi-turn, multi-tool interactions across environments.

\section*{Limitations}
In this paper, we propose EnvScaler for automated synthesis of tool-interactive environments. However, there are still some limitations, as follows:
(1) Although LLMs are not directly used as environment simulators, the entire construction process relies on LLM synthesis, which may introduce biases compared to real systems (e.g., business logic, state definitions, and task design).  
(2) In terms of domain coverage, EnvScaler mainly targets domain-specific, stateful environments, with limited support for open environments such as web search or information access.  
(3) In terms of feature modeling, EnvScaler focuses on tool–state interactions but lacks explicit simulation of real-system characteristics such as interface latency, network fluctuations, and error patterns.
(4) In terms of modality, EnvScaler only supports text-based tool inputs and outputs, without incorporating multimodal tools involving images, audio, or other modalities.
(5) In terms of scale, although EnvScaler is highly scalable, the current study is limited to around 200 synthesized environments and 7K scenarios due to experimental resources.

\section*{Acknowledgments}
This work was supported by National Natural Science Foundation of China No. 62272467 and No. 625B2178. The work was partially done at the Beijing Key Laboratory of Research on Large Models and Intelligent Governance.

\bibliography{custom}

\begin{figure*}[!t]
    \centering
    \resizebox{0.85\textwidth}{!}{
    \includegraphics{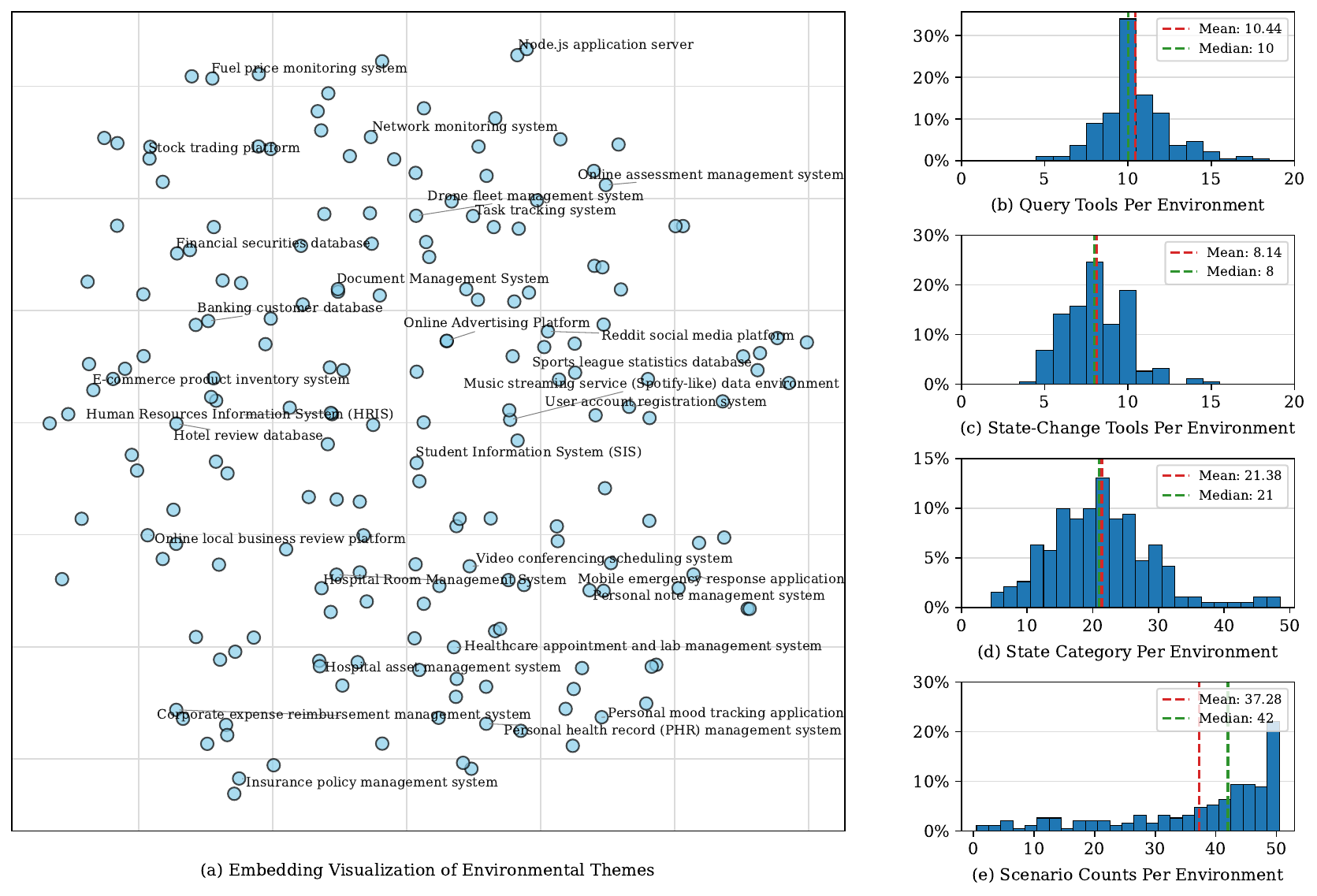}}
    \caption{Diversity and statistical distributions of 191 synthesized environments.}
    \label{fig:env_topic_embedding}
\end{figure*}

\appendix

\section{Details of EnvScaler}
\label{app:env_synth_details}

\subsection{Details of Prompts for SkelBuilder}
\label{subapp:env_prompt}
\begin{itemize}
[itemsep=4pt,topsep=0pt,parsep=0pt,leftmargin=9pt]
\item Figures \ref{fig:task_judge_prompt} and \ref{fig:infer_env_topic_prompt} show the prompts for task filtering and environment‑theme inference in the Task‑driven Environment Discovery stage.
\item Figures \ref{fig:infer_state_rules_prompt} and \ref{fig:infer_operation_prompt} present the prompts used to plan environment states, rules, and operations during the environment‑modeling stage, while Figures \ref{fig:convert_state_to_class_prompt} and \ref{fig:convert_op_to_func_prompt} show the prompts for converting these blueprints into executable programs.
\item Figures \ref{fig:testing_agent_prompt} and \ref{fig:checking_agent_prompt} show the system prompts for the testing agent and checking agent in the Dual‑Agent Env Assessment stage, respectively.
\end{itemize}

\subsection{Details of Prompts for ScenGenerator}
\label{subapp:scenario_prompt}
Figures \ref{fig:gen_init_state_prompt} and \ref{fig:gen_task_prompt} show the prompts for generating the environment’s initial state and tasks, respectively. 
Figures \ref{fig:gen_checklist_prompt} and \ref{fig:gen_check_func_prompt} show the prompts for decomposing tasks into checklists and converting each check item into a verification function.

\subsection{Details of Prompts for LLM Agents}
\label{subapp:llm_agent_prompt}
Figures \ref{fig:agent_non_conv_prompt} and \ref{fig:agent_conv_prompt} show the system prompts guiding LLMs complete tasks under the \textit{Non‑conversation} and \textit{Conversation} settings.
Figure \ref{fig:sys_prompt_user} shows the system prompt for guiding a LLM to act as a user.

\subsection{Details and Statistics of 191 Synthesized Environment}
\label{subapp:env_statistics}
Figure~\ref{fig:env_topic_embedding} shows detailed statistics for the 191 synthesized environments.
By embedding each environment’s theme description and applying t‑SNE, we obtain Figure~\ref{fig:env_topic_embedding}(a), where the point dispersion demonstrates the diversity of environment themes.
Figures~\ref{fig:env_topic_embedding}(b) and (c) show the distribution of tool counts across environments, while Figure~\ref{fig:env_topic_embedding}(d) shows the distribution of state categories.
Figure~\ref{fig:env_topic_embedding}(e) shows synthesized scenario counts per environment after removing unqualified scenario.

\subsection{Example of Synthesized Environment}
\label{subapp:env_example}
In Example \ref{box:program-example}, we provide a program example of synthesized environments,  and Table~\ref{tab:tool_schema_example} shows the corresponding tool interfaces.

\subsection{Example of Environment's Task Scenario}
\label{subapp:scenario_example}
\begin{itemize}
[itemsep=4pt,topsep=0pt,parsep=0pt,leftmargin=9pt]
\item \textbf{Initial State Configuration:} Figure~\ref{fig:init_config_example} shows an example of an initial state configuration, which is used to initialize the environment’s state data.
\item \textbf{Task:} Figure~\ref{fig:task_example} presents an example task under this initial state.
\item \textbf{State Check Functions:} Figure~\ref{tab:check_func_example} shows the checklist associated with the task, along with the check function for each checkpoint, which are used to compute a trajectory’s reward score.
\end{itemize}

\subsection{Example Trajectory of LLM Interacting with Synthesized Environment}
\label{subapp:traj_example}
\begin{itemize}
[itemsep=4pt,topsep=0pt,parsep=0pt,leftmargin=9pt]
\item \textbf{Non-Conversation:} Table~\ref{tab:non_conv_traj_example} shows a trajectory example of Qwen3‑30B-A3B‑Thinking‑2507 with the environment under Non‑Conv setting.
\item \textbf{Conversation:} Table~\ref{tab:conv_traj_example} shows a trajectory example of Qwen3‑30B‑A3B‑Thinking‑2507 under Conversation setting. Both trajectories correspond to the same task scenario.
\end{itemize}

\section{Details of Experiments}
\label{app:experiment_details}

\subsection{Details of Evaluation Benchmarks}
\label{subapp:dataset_info}
\begin{itemize}
[itemsep=4pt,topsep=0pt,parsep=0pt,leftmargin=9pt]
\item 
\textbf{BFCL-v3 Multi-Turn}~\citep{patil2025the}:
The BFCL‑v3 Multi‑Turn dataset evaluates the ability of LLMs to consistently and accurately perform function calls across multi‑turn dialogues, covering 8 environments including vehicle control, transactions, ticketing, and file systems. In addition to the \textbf{Base} subset for standard tasks, it includes \textbf{Missing Parameters}, \textbf{Missing Functions}, and \textbf{Long Context} subsets to test handling of incomplete information, unavailable functions, and information‑dense long contexts. The evaluation applies both state checking and response checking as dual criteria.
\item 
\textbf{Tau-Bench}~\citep{yao2025taubench}: 
Tau-Bench is a benchmark for evaluating LLM agents in realistic human–AI interaction scenarios, focusing on their ability to engage in multi‑turn dialogues, invoke domain‑specific APIs, and follow complex business rules.
It includes two subsets: 
(1) \textbf{Retail}: Simulated customer service for retail, such as order cancellation/modification, returns and exchanges.
(2) \textbf{Airline}: Simulated airline customer service, such as ticket booking, rescheduling, baggage and insurance handling, with more complex rules and greater reasoning challenges.
\item 
\textbf{ACEBench-Agent}~\citep{chen-etal-2025-acebench}:
It evaluates LLMs’ ability to perform multi-turn and multi-step tool calls in dynamic, real-world environments, covering scenarios such as mobile apps, food delivery platforms, financial services, and travel booking. It consists of two subsets:
(1) \textbf{Multi-turn}: The user interacts with the LLM multiple times during the task, gradually providing information or adjusting requirements.
(2) \textbf{Multi-step}: The user gives the task only once at the start, and the LLM autonomously calls tools until it decides the task is complete.
\end{itemize}
Table~\ref{tab:eval_dataset_info} present the statistics of each benchmark.

\begin{table}[t]
\small
\centering
\setlength{\tabcolsep}{1mm}
    \begin{tabular}{lccc}
    \toprule
        \textbf{Dataset} & \textbf{\# Env} & \textbf{\# Tools Per Env} & \textbf{\# Task} \\
        \midrule
        \textbf{BFCL-v3 Multi-Turn} & \textbf{8} & \textbf{16.13} & \textbf{800} \\
        \quad - Base & 8 & 16.13 & 200 \\
        \quad - Miss-Param & 8 & 16.13 & 200 \\
        \quad - Miss-Func & 8 & 16.13 & 200 \\
        \quad - Long-Context & 8 & 16.13 & 200 \\
        \midrule
        \textbf{Tau-Bench} & \textbf{2} & \textbf{14} & \textbf{175} \\
        \quad - Retail & 1 & 15 & 115 \\
        \quad - Airline & 1 & 13 & 50 \\
        \midrule
        \textbf{ACEBench-Agent (EN)} & \textbf{4} & \textbf{7.75} & \textbf{50} \\
        \quad - Multi-Step & 3 & 6 & 20 \\
        \quad - Multi-Turn & 4 & 7.75 & 30 \\
    \bottomrule
    \end{tabular}
\caption{Statistics of evaluation benchmarks.}
\label{tab:eval_dataset_info}
\end{table}

\subsection{Details of Implementation}
\label{subapp:implement_details}

\begin{itemize}
[itemsep=4pt,topsep=0pt,parsep=0pt,leftmargin=9pt]
\item 
\textbf{SFT. }
We use Qwen3-30B-A3B-Thinking-2507 as the teacher model to interact with 140 environments under two interaction settings. We remove trajectories that contain invalid formats or are judged by the LLM as impossible to complete, resulting in about 9K final trajectories.
For SFT, We use the LlamaFactory framework~\citep{zheng2024llamafactory}.
In thinking mode, Qwen3 automatically removes all rounds’ reasoning process when applying the \texttt{chat\_template}. To let the model learn the reasoning in each round, we split an $n$‑round sample into $n$ sub-samples (corresponding rounds are 1,2,...,$n$). Only the final round’s reasoning and action in each sub-sample are supervised and optimized (implemented via the \texttt{mask\_history} hyperparameter).
We train for 3 epochs with a learning rate of 1e‑6, a maximum sequence length of 32K tokens, and an effective batch size of 256 after gradient accumulation.
\item 
\textbf{RL. }
We conduct RL using the Reinforce++ algorithm~\citep{hu2025reinforcestabilizingcriticfreepolicy} within the ROLL framework~\citep{wang2025reinforcement}. We retain the KL constraint with a coefficient of 0.1 and use a learning rate of $1.0\times10^{-6}$. In each training step, we sample 64 tasks and rollout 8 trajectories per task in the Non-Conversation setting, for a total of up to 200 training steps. The maximum trajectory length is set to 32K tokens, and the maximum generation length per step is capped at 4K tokens.
\item
\textbf{Evaluation. }
We evaluate with a temperature of 0.7 and report the average over three runs. Following the Qwen3 setup, we keep only historical actions and remove historical reasoning process.
For both baselines and our trained models, we use each model’s own function-calling interface to ensure consistent tool-use formats.
Because ACEBench uses the \texttt{[func\_name(param)]} prompt format by default, we modify the official code to support LLMs’ native function-calling interface.
For Tau-Bench, we use GPT-4.1 in ReAct mode to simulate users.  
For the Multi-Turn subset of ACEBench-Agent, we likewise use GPT-4.1 for user simulation.
We set the maximum context length to 64K to match the Long-Context subset of BFCL-v3.
\end{itemize}

\section{More Experiments}
\label{app:more_experiments}

\begin{table}[t]
\small
\centering
\setlength{\tabcolsep}{0.8mm}
\begin{tabular}{lccc}
\toprule
   \textbf{\makecell{Model\\(Non-Thinking)}} & \textbf{\makecell{BFCL\\Multi-Turn}} & \textbf{\makecell{ACEBench\\-Agent}} & \textbf{\makecell{Tau-\\Bench}} \\
   \midrule
    Qwen3-4B & 6.88 & 30.56 & 20.15 \\ 
    \quad+ APIGen & $\text{6.75}_{\textcolor{deepred}{\text{-0.13}}}$ 
                  & $\text{20.84}_{\textcolor{deepred}{\text{-9.72}}}$ 
                  & $\text{14.71}_{\textcolor{deepred}{\text{-5.44}}}$ \\
    \quad+ APIGen-MT & $\text{3.00}_{\textcolor{deepred}{\text{-3.88}}}$ 
                     & $\text{32.50}_{\textcolor{deepgreen}{\text{+1.94}}}$ 
                     & $\text{33.93}_{\textcolor{deepgreen}{\text{+13.78}}}$ \\
    \quad+ EnvScaler (SFT) & $\text{17.88}_{\textcolor{deepgreen}{\text{+11.00}}}$ 
                          & $\text{41.67}_{\textcolor{deepgreen}{\text{+11.11}}}$ 
                          & $\text{18.57}_{\textcolor{deepred}{\text{-1.58}}}$ \\
\bottomrule
\end{tabular}
\caption{Performance of Qwen3-4B after training on different datasets in non-thinking mode.}
\label{tab:non_think_result}
\end{table}

\subsection{Results of Training in Non-thinking Mode}
\label{subapp: non_think_result}
Table~\ref{tab:non_think_result} shows the performance of Qwen3‑4B  training on different SFT datasets in non‑thinking mode. We also compare the APIGen~\citep{liu2024apigen} and APIGen‑MT~\citep{prabhakar2025apigen} datasets because they only provide non‑thinking samples. APIGen contains 60K single‑turn tool‑use examples; APIGen‑MT contains 5K multi‑turn examples, but they cover only two environments, which are identical to those in Tau-Bench. For EnvScaler, we remove the teacher LLM’s reasoning process and keep only the action sequences. 
First, APIGen leads to degraded performance on all benchmarks, indicating that relying solely on single‑turn supervision cannot improve an LLM’s multi‑turn tool‑use capabilities. Due to the environments being the same as in Tau-Bench, APIGen‑MT provides a significant boost on Tau-Bench, but it fails to transfer to BFCL‑MT or ACEBench‑Agent. Moreover, since this dataset uses only the Conversation setting, the model tends to talk to the user before using tools, which conflicts with the style of BFCL‑MT. This highlights the importance of training in both settings. Finally, EnvScaler (SFT) brings significant improvements on BFCL‑MT and ACEBench‑Agent but slightly degrades performance on Tau-Bench. We believe this is because Tau-Bench relies more on reasoning, 
while in non‑thinking mode the LLM learns only actions and not the reasoning chain, making it difficult to produce correct actions during evaluation.

\subsection{Analysis of Discarded Envs}

\begin{figure}[!t]
    \centering
    \resizebox{0.48\textwidth}{!}{
    \includegraphics{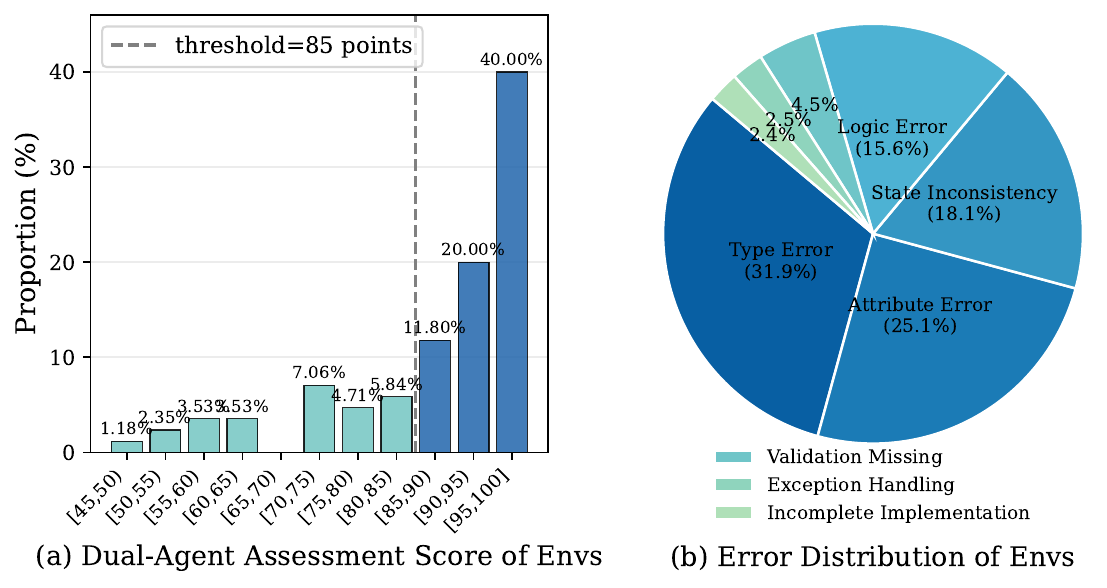}}
    \caption{Score and error distribution of environments in dual-agent loop assessment.}
    \label{fig:env_error_analysis}
\end{figure}

As stated in Section \ref{subsec:env_practice}, we conducted 100 rounds of cyclic testing for each synthetic environment and retained only those with a test success rate above 0.85. From 266 synthetic environments, we selected 191, corresponding to a rejection rate of 28.2\%.
In Figure~\ref{fig:env_error_analysis} (a), we present the distribution of scores for all environments in the dual-agent loop assessment.
To systematically analyze the reasons for environment failures, we extract all test cases identified as failures from the logs in the dual-agent loop quality inspection. We use Claude-4.5-Sonnet to classify these errors. As shown in Figure~\ref{fig:env_error_analysis} (b), the primary issue is type errors or attribute errors, followed by the consistency of state management.

\appendix

\section{Use of AI Assistant}
In this paper, we only use ChatGPT\footnote{\url{https://chatgpt.com/}} for grammar proofreading and spell checking.

\input{env_skel_synth_prompt}

\input{env_scen_synth_prompt}
\input{llm_agent_prompt}

\clearpage
\onecolumn
\refstepcounter{programbox}\label{box:program-example}
\begin{tcolorbox}[
    enhanced jigsaw,
    colback=white!10!white,
    colframe=black!75!black,
    left=0.5pt, right=0.5pt,
    boxsep=1pt,
    title={Example~\theprogrambox: Program Example of Synthesized Environment},
    breakable
]
    \vspace{-0.3cm}
\begin{lstlisting}[style=python]
from typing import Dict, List, TypedDict
import uuid
import time

class UserInfo(TypedDict):
    _id: str
    phone_number: str
    name: str
    profile_info: str

class ContactInfo(TypedDict):
    _id: str
    contact_id: str
    contact_phone_number: str
    contact_name: str

class MessageInfo(TypedDict):
    message_id: str
    sender_id: str
    receiver_id: str
    receiver_phone_number: str
    content: str
    timestamp: str
    delivery_status: str
    read_status: str

class ConversationInfo(TypedDict):
    conversation_id: str
    user_ids: List[str]
    message_id: str

class MobileMessagingApplication:
    def __init__(self, config: dict):
        self.users: Dict[str, UserInfo] = {}
        self.contacts: Dict[str, ContactInfo] = {}
        self.messages: Dict[str, MessageInfo] = {}
        self.conversations: Dict[str, ConversationInfo] = {}

        # Constraints:
        # - Only registered users can send messages
        # - Phone numbers must be valid before sending messages
        # - Message delivery status must be updated (e.g., sent, delivered, failed)
        # - Conversation history is retained for each user-contact pair
        # - Duplicate messages to the same phone number are allowed but tracked separately

        self.config = config

    def get_user_by_phone_number(self, phone_number: str) -> dict:
        """Look up a user by their phone number.
        Constraints:
            - Only registered users are present in the system.
            - Phone number must match a user entry; otherwise, return error."""
        for user in self.users.values():
            if user["phone_number"] == phone_number:
                return {"success": True, "data": user}
        return {"success": False, "error": "User with this phone number not found"}

    def get_user_by_id(self, user_id: str) -> dict:
        """Retrieve user information by user ID.
        Constraints:
            - user_id must correspond to an existing registered user."""
        user = self.users.get(user_id)
        if user is None:
            return { "success": False, "error": "User not found" }
        return { "success": True, "data": user }

    def list_all_users(self) -> dict:
        """List all registered user accounts.
        Constraints:
            - Returns the current set of registered users.
            - No failure unless system corruption (not handled here)."""
        user_list = list(self.users.values())
        return { "success": True, "data": user_list }

    def get_contact_by_phone_number(self, phone_number: str) -> dict:
        """Retrieve contact information using a contact's phone number.
        Constraints:
            - The provided phone number must correspond to an existing contact."""
        for contact in self.contacts.values():
            if contact["contact_phone_number"] == phone_number:
                return {"success": True, "data": contact}
        return {"success": False, "error": "No contact found with this phone number."}

    def get_contact_by_id(self, contact_id: str) -> dict:
        """Look up and return contact info by contact ID.
        Constraints:
            - Contact must exist in the system."""
        contact = self.contacts.get(contact_id)
        if contact is None:
            return { "success": False, "error": "Contact not found" }
        return { "success": True, "data": contact }

    def list_user_contacts(self, user_id: str) -> dict:
        """Lists all contacts associated with a particular user.
        Constraints:
            - The provided user_id must correspond to a registered user."""
        if user_id not in self.users:
            return {"success": False, "error": "User does not exist"}
        contacts = [
            contact_info for contact_info in self.contacts.values()
            if contact_info.get("contact_id") == user_id
        ]
        return {"success": True, "data": contacts}

    def validate_phone_number(self, phone_number: str) -> dict:
        """Check if a phone number is valid for messaging.
        Constraints:
            - Phone number should be a non-empty string.
            - Phone number should only contain digits, possibly beginning with '+'.
            - Length must be between 10 and 15 (excluding '+')."""
        if not isinstance(phone_number, str):
            return { "success": False, "error": "Phone number must be a string" }
        stripped = phone_number.lstrip('+')
        if not stripped.isdigit():
            return {
                "success": True,
                "data": {"valid": False, "reason": "Phone number must contain only digits after optional '+'"}
            }
        if len(stripped) < 10 or len(stripped) > 15:
            return {
                "success": True,
                "data": {"valid": False, "reason": "Phone number length must be between 10 and 15 digits"}
            }
        # Passed all checks
        return {
            "success": True,
            "data": {"valid": True, "reason": "Phone number is valid for messaging"}
        }

    def get_messages_by_phone_number(self, phone_number: str) -> dict:
        """Retrieve all messages (sent or received) associated with the specified phone number.
        Constraints:
            - Matches both messages where receiver_phone_number == phone_number,
              and messages sent by users whose phone_number == phone_number.
            - Returns empty list if no matches found; this is not considered an error.
            - Input phone number should be a string."""
        if not isinstance(phone_number, str):
            return { "success": False, "error": "Invalid phone number format" }
        # Find all user IDs whose phone number matches the input (for sent messages)
        sender_ids = [
            user_id for user_id, user_info in self.users.items()
            if user_info["phone_number"] == phone_number
        ]
        result = [
            msg for msg in self.messages.values()
            if ( msg["receiver_phone_number"] == phone_number or msg["sender_id"] in sender_ids)]

        return { "success": True, "data": result }

    def get_conversation_by_user_and_contact(self, user_id: str, contact_id: str) -> dict:
        """Get the conversation history (messages) between a user and a given contact.
        Constraints:
            - User id must exist in the platform.
            - Contact id must exist in contacts.
            - Returns all messages in the user-contact conversation(s).
        """
        if user_id not in self.users:
            return { "success": False, "error": "User not found" }
        if contact_id not in self.contacts:
            return { "success": False, "error": "Contact not found" }
        # Find relevant conversations:
        conversations = [
            convo for convo in self.conversations.values()
            if user_id in convo["user_ids"] and contact_id in convo["user_ids"]
        ]
        if not conversations:
            return { "success": False, "error": "No conversation found between user and contact" }
        # Gather all related message infos:
        messages = []
        for convo in conversations:
            msg_id = convo["message_id"]
            if msg_id in self.messages:
                messages.append(self.messages[msg_id])

        return { "success": True, "data": messages }

    def get_message_by_id(self, message_id: str) -> dict:
        """Retrieve information for a specific message given its unique message ID.
        Constraints:
            - The message ID must exist in the environment's messages."""
        message = self.messages.get(message_id)
        if not message:
            return { "success": False, "error": "Message not found" }
        return { "success": True, "data": message }

    def get_message_delivery_status(self, message_id: str) -> dict:
        """Check and return the current delivery status of a specific message.
        Constraints:
            - message_id must exist in the system."""
        msg = self.messages.get(message_id)
        if not msg:
            return { "success": False, "error": "Message ID not found" }
        return {
            "success": True, "data": {"message_id": message_id, "delivery_status": msg["delivery_status"]}}

    def send_message(self, sender_id: str, receiver_phone_number: str, content: str) -> dict:
        """Create and send a new message from a registered user to a valid phone number.
        Constraints:
            - Only registered users can send messages.
            - Phone numbers must be valid before sending.
            - Message delivery status must be set ('sent').
            - Conversation history must include this message.
            - Content must not be empty."""
        # Sender must be a registered user
        if sender_id not in self.users:
            return {"success": False, "error": "Sender is not a registered user."}
        # Content must not be empty
        if not content or content.strip() == "":
            return {"success": False, "error": "Message content cannot be empty."}
        # Validate phone number
        if not hasattr(self, "validate_phone_number") or not callable(self.validate_phone_number):
            return {"success": False, "error": "Phone number validation method is unavailable."}
        valid_check = self.validate_phone_number(receiver_phone_number)
        if not valid_check.get("success", False):
            return {"success": False, "error": "Receiver phone number is invalid."}

        # Determine receiver user (if exists)
        receiver_id = None
        for uid, user in self.users.items():
            if user["phone_number"] == receiver_phone_number:
                receiver_id = uid
                break
        # Create unique message_id (use combination of sender, receiver, timestamp, and an increment)
        message_id = str(uuid.uuid4())
        timestamp = str(time.time())
        # Initial delivery status and read status
        delivery_status = "sent"
        read_status = "unread"
        # Create message entry
        message_info = {
            "message_id": message_id,
            "sender_id": sender_id,
            "receiver_id": receiver_id if receiver_id else "",
            "receiver_phone_number": receiver_phone_number,
            "content": content,
            "timestamp": timestamp,
            "delivery_status": delivery_status,
            "read_status": read_status
        }
        self.messages[message_id] = message_info

        # Maintain conversation history: 
        # Find if a conversation exists between sender and receiver (user_id or phone_number)
        # Use sorted user_ids (if possible), else pseudo user_id from phone number
        user_ids = [sender_id]
        if receiver_id:
            user_ids.append(receiver_id)
        else:
            # Create pseudo id for phone number to track conversation
            user_ids.append(f"phone:{receiver_phone_number}")
        user_ids_sorted = sorted(user_ids)
        # Try to find existing conversation
        conversation_id = None
        for cid, conv in self.conversations.items():
            if sorted(conv['user_ids']) == user_ids_sorted:
                conversation_id = cid
                break
        if conversation_id is None:
            # Create a new conversation
            conversation_id = str(uuid.uuid4())
            self.conversations[conversation_id] = {
                "conversation_id": conversation_id,
                "user_ids": user_ids_sorted,
                "message_id": message_id
            }
        else:
            # Update conversation: for simplicity, replace message_id (could be made a list in real impl.)
            # To keep track, we simply overwrite the latest message_id (history is in messages)
            self.conversations[conversation_id]["message_id"] = message_id

        return {"success": True, "message": "Message sent successfully."}

    def update_message_delivery_status(self, message_id: str, new_status: str) -> dict:
        """Set or update the delivery status for a particular message.
        Constraints:
            - The message_id must exist.
            - Optionally, enforce that new_status is among allowed values ('sent', 'delivered', 'failed')."""
        allowed_statuses = {"sent", "delivered", "failed"}
        if message_id not in self.messages:
            return { "success": False, "error": "Message not found" }
        if new_status not in allowed_statuses:
            return { "success": False, "error": f"Invalid delivery status: {new_status}" }
        self.messages[message_id]["delivery_status"] = new_status
        return { "success": True, "message": f"Delivery status updated for message {message_id}" }

    def create_contact(self, contact_id: str, contact_phone_number: str, contact_name: str) -> dict:
        """Add a new contact to the specified user's contact list.
        Constraints:
            - Only registered users can add contacts.
            - Phone numbers must be valid before creating a contact.
            - Duplicate contacts (same phone number) are allowed but tracked separately."""
        # Check that the user exists
        if contact_id not in self.users:
            return { "success": False, "error": "User does not exist" }
        # Validate phone number using the environment's function
        if not hasattr(self, "validate_phone_number"):
            return { "success": False, "error": "Phone number validation not available" }
        validation_result = self.validate_phone_number(contact_phone_number)
        if not (isinstance(validation_result, dict) and validation_result.get("success")):
            return { "success": False, "error": "Phone number is invalid" }
        # Generate new contact _id (ensure uniqueness)
        new_contact_id = f"contact_{len(self.contacts)+1}"
        # Create and store the contact
        contact_info = {
            "_id": new_contact_id,
            "contact_id": contact_id,
            "contact_phone_number": contact_phone_number,
            "contact_name": contact_name
        }
        self.contacts[new_contact_id] = contact_info

        return { "success": True, "message": "Contact created" }

    def add_message_to_conversation(self, conversation_id: str, message_id: str) -> dict:
        """Insert a new message into the conversation history for the appropriate user-contact pair.
        Constraints:
            - Both the conversation and the message must exist.
            - Each message can be tracked in the history; duplicate messages to same phone are allowed but tracked separately.
            - User IDs in message context should match conversation participants (not strictly enforced)."""
        # Check conversation existence
        if conversation_id not in self.conversations:
            return { "success": False, "error": "Conversation does not exist." }
        if message_id not in self.messages:
            return { "success": False, "error": "Message does not exist." }
        # Get conversation and message info
        conv_info = self.conversations[conversation_id]
        msg_info = self.messages[message_id]
        # Optionally check participants (message sender/receiver in conversation user_ids)
        relevant_user_ids = conv_info.get("user_ids", [])
        msg_participants = [msg_info["sender_id"], msg_info["receiver_id"]]
        missing = [uid for uid in msg_participants if uid not in relevant_user_ids]
        # Not strictly enforced per constraints, but warn if mismatch.
        # For this environment, conversation history is retained for each user/contact pair.
        # Since ConversationInfo maps one message_id, we append a new entry with same conv_id and user_ids
        new_convo_entry = {
            "conversation_id": conversation_id,
            "user_ids": relevant_user_ids,
            "message_id": message_id
        }
        # Insert new entry into conversations dict with a unique key
        # Use 'conversation_id:message_id' as a compound key to allow duplicate messages
        convo_key = f"{conversation_id}:{message_id}"
        self.conversations[convo_key] = new_convo_entry

        return { "success": True, "message": "Message added to conversation history." }

    def create_conversation(self, user_id1: str, user_id2: str) -> dict:
        """Initiate a new conversation record between two users or a user and a contact, if such a conversation does not already exist.
        Constraints:
            - Both participants must exist (in users or contacts).
            - Conversation between same pair must be unique.
            - Participants must not be identical."""
        # Check that both participants exist in the system
        valid_ids = set(self.users.keys()) | set(self.contacts.keys())
        if user_id1 not in valid_ids:
            return { "success": False, "error": f"Participant 1 ({user_id1}) does not exist." }
        if user_id2 not in valid_ids:
            return { "success": False, "error": f"Participant 2 ({user_id2}) does not exist." }
        if user_id1 == user_id2:
            return { "success": False, "error": "Cannot create conversation with oneself." }

        # Check if a conversation between these two IDs already exists
        for conv in self.conversations.values():
            ids_set = set(conv["user_ids"])
            if ids_set == set([user_id1, user_id2]):
                return {
                    "success": False,
                    "error": "Conversation between these participants already exists.",
                    "conversation_id": conv["conversation_id"]
                }
        conversation_id = str(uuid.uuid4())
        conversation_info: ConversationInfo = {
            "conversation_id": conversation_id,
            "user_ids": [user_id1, user_id2],
            "message_id": ""  # Initial, no message yet
        }
        self.conversations[conversation_id] = conversation_info
        return {
            "success": True,
            "message": "Conversation created",
            "conversation": conversation_info
        }

    def mark_message_as_read(self, message_id: str) -> dict:
        """Mark a message as read by updating its read_status field.
        Constraints:
            - message_id must exist in the system.
            - Operation is idempotent; if already 'read', report success."""
        message = self.messages.get(message_id)
        if not message:
            return { "success": False, "error": "Message not found." }
        if message["read_status"] == "read":
            # Already marked as read, treat as success
            return { "success": True, "message": "Message marked as read." }
        # Mark as read
        message["read_status"] = "read"
        return { "success": True, "message": "Message marked as read." }

    def delete_message(self, message_id: str) -> dict:
        """Remove a message from the system. If message exists, it's deleted and any references
        in conversations are also removed.
        Constraints:
        - Message must exist in the system.
        - References to this message in conversation objects are also removed for integrity."""
        if message_id not in self.messages:
            return {"success": False, "error": f"Message {message_id} does not exist."}
        # Remove references in conversations
        for conv in self.conversations.values():
            if conv["message_id"] == message_id:
                conv["message_id"] = ""  # Remove reference (could also delete the conversation if message is sole content)
                # Alternatively, could maintain a messages list if the schema changes
        # Remove the message itself
        del self.messages[message_id]
        return {
            "success": True,
            "message": f"Message {message_id} has been deleted."
        }

    def archive_conversation(self, user_id: str, conversation_id: str) -> dict:
        """Archive or hide a conversation for a specific user.
        Constraints:
            - Only registered users may archive conversations.
            - Conversation must exist.
            - User must be a participant in the conversation.
            - Archiving is idempotent (archiving again is still success)."""
        # Ensure tracking exists
        if not hasattr(self, "archived_conversations"):
            self.archived_conversations: Dict[str, List[str]] = {}
        if user_id not in self.users:
            return { "success": False, "error": "User does not exist." }
        if conversation_id not in self.conversations:
            return { "success": False, "error": "Conversation does not exist." }
        conversation_info = self.conversations[conversation_id]
        if user_id not in conversation_info["user_ids"]:
            return { "success": False, "error": "User is not a participant in the conversation." }
        archived_list = self.archived_conversations.setdefault(user_id, [])
        if conversation_id not in archived_list:
            archived_list.append(conversation_id)

        return { "success": True, "message": "Conversation archived for user." }

    def delete_conversation(self, conversation_id: str, requestor_id: str) -> dict:
        """Remove a conversation from history if the action is allowed.
        Constraints:
            - Only admins or users who are part of the conversation can delete it.
            - If conversation does not exist, fail.
            - Conversation removal does not delete referenced messages."""
        # Check if conversation exists
        conv = self.conversations.get(conversation_id)
        if not conv:
            return { "success": False, "error": "Conversation does not exist." }
        # Check requestor existence
        user_info = self.users.get(requestor_id)
        if not user_info:
            return { "success": False, "error": "Requestor does not exist." }
        # Permission check: Admin or participant
        is_admin = (user_info.get("profile_info", "") == "admin")
        is_participant = requestor_id in conv.get("user_ids", [])
        if not (is_admin or is_participant):
            return { "success": False, "error": "Permission denied." }
        # Delete the conversation
        del self.conversations[conversation_id]

        return { "success": True, "message": "Conversation deleted successfully." }
\end{lstlisting}
    \vspace{-0.3cm}
\end{tcolorbox}
\input{tool_schema_example}
\twocolumn
\clearpage
\input{task_exmaple}
\input{traj_example}

\end{document}

%% file: env_skel_synth_prompt.tex
\begin{figure*}
\begin{tcolorbox}[
colback=white!10!white,
colframe=black!75!black,
left=0.5pt, right=0.5pt,
boxsep=1pt,
title=SkelBuilder: Prompt for Task Filtering,
]
\vspace{-0.3cm}
\begin{lstlisting}
You are a system that filters natural language tasks to determine if they are state-dependent, actionable requests within a persistent, domain-specific environment.

### Core Definition
We are ONLY looking for tasks that meet all of the following criteria:
1. Persistent Environment 
   - The query is about a domain where:
   - There is a live, ongoing state that can be read or changed
   - The environment supports both: a) Information queries about current state (read operations)  b) Explicit state-changing actions (create, update, delete, move, cancel, etc.)
2. State Dependency 
   - The task cannot be answered correctly without:
   - Inspecting the actual current data or configuration in the environment, and/or
   - Executing an operation that modifies that data.
3. Domain Specificity
   - The environment is not general-purpose knowledge; it is a structured system such as: File management system with stored files/folders, Calendar/scheduling system, Other specialized platforms with records that persist over time
4. Actionability in Context
   - The query must correspond to an actionable operation or status check within the actual environment (not hypothetical).

### Eligible Task Types
- State queries: "Is invoice #1024 paid?" / "What meetings are scheduled for Wednesday?"
- State modification operations: "Upload the proposal.pdf to the project folder" / "Cancel order #4512" / "Move meeting to 3 PM"

### Explicit Exclusions
A request is NOT eligible if it is: Open-domain factual Q&A, Casual conversation, Content creation, Isolated reasoning or calculations without accessing persisted state

### Judgment Rule - Be strict:
Choose YES only if:
- The query cannot be answered from general knowledge alone
- AND it requires real-time access to persistent state in a domain-specific environment
- AND it targets an actionable operation (either a read or a write to that environment)
- AND the environment has the capability for both queries and modifications
If any criterion is missing - NO.
\end{lstlisting}
\vspace{-0.3cm}
\end{tcolorbox}
\caption{The prompt for filtering tasks situated within a domain-specific, stateful environment.}
\label{fig:task_judge_prompt}
\end{figure*}

\begin{figure*}
\begin{tcolorbox}[
colback=white!10!white,
colframe=black!75!black,
left=0.5pt, right=0.5pt,
boxsep=1pt,
title=SkelBuilder: Prompt for Inferring Env Description,
]
\vspace{-0.3cm}
\begin{lstlisting}
You are a Task Analyst. Given a raw task description, your objective is to identify the most plausible stateful and domain-specific environment in which this task would naturally occur.  

The chosen environment should strike a balance: not so broad as to be meaningless, and not so narrow as to apply only to a single, highly specific case. It should be scoped such that this task, along with similar related tasks, can be executed meaningfully. If multiple environments seem equally plausible, select one at random rather than listing all possibilities.  

Your response must include the following sections:  

1. # Analysis  
   - Explain the reasoning process used to connect the task to the chosen environment.  
   - Note any relevant entities, constraints, relationships, or dynamics implied by the task.  

2. # Environment Summary  
   - Provide a concise label for the environment type.  

3. # Environment Introduction  
   - Introduce the environment itself, without referring to the current task.  
   - Focus on its inherent structure, the nature of the state it maintains, typical operations it supports, and its general scope in real-world usage.  
   - Limit to approximately three sentences.  

4. # Metrics  
   - Usefulness: 1-10  
     Reflects how broadly applicable and valuable this environment is in real-world scenarios. Higher scores indicate environments relevant to many contexts and industries.  
   - Modelability: 1-10  
     Indicates how straightforward it would be to represent this environment using a single Python class, with attributes holding state and methods performing reading, writing, and querying operations. Higher scores indicate simpler, more self-contained structures.  
\end{lstlisting}
\vspace{-0.3cm}
\end{tcolorbox}
\caption{The prompt for inferring environment description from existing tasks.}
\label{fig:infer_env_topic_prompt}
\end{figure*}

\begin{figure*}
\begin{tcolorbox}[
colback=white!10!white,
colframe=black!75!black,
left=0.5pt, right=0.5pt,
boxsep=1pt,
title=SkelBuilder: Prompt for Planning States \& Rules,
]
\vspace{-0.3cm}
\begin{lstlisting}
You are an expert task and environment analyst. Given an environment description and a example task in this environment, infer the set of state variables (state space) that the environment maintained.  
The state should not be too broad (e.g. "all possible data in an e-commerce system"), nor too narrow (only for this single task).  Instead, reasonably design it to support this task and similar tasks in the same environment.  
Your output must follow the format below (do not include any other text):
# Analysis
[Your thought process: What states are involved in the environment? What entities/attributes are needed to be tracked? What constraints or rules exist in the environment? ...]
# State Space Definition
- Entity: EntityName1  
  - Attributes: Attribute1, Attribute2, ...
  - Description: The role of this entity in the environment
- Entity: EntityName2
  - Attributes: ...
  - Description: ...
# Constraints & Rules
- Constraint 1
- Constraint 2
...
\end{lstlisting}
\vspace{-0.3cm}
\end{tcolorbox}
\caption{The prompt for planning state and rules of the environment.}
\label{fig:infer_state_rules_prompt}
\end{figure*}

\begin{figure*}
\begin{tcolorbox}[
colback=white!10!white,
colframe=black!75!black,
left=0.5pt, right=0.5pt,
boxsep=1pt,
title=SkelBuilder: Prompt for Planning Tools,
]
\vspace{-0.3cm}
\begin{lstlisting}
You are an expert in building and analyzing agent environments.
Given an environment summary, introduction, state space definition, constraint rules, Python base class definition, and example task, your goal is to analyze the current environment and then generate the list of operations needed to support the task in this environment (including information query class and state modification class).
Each operation will be converted into a class function for the Agent to use in subsequent steps.

Key Points:  
- Operations are divided into 2 categories: Information Query Class and State Change Class.  
- Each operation includes: operation name + brief description.  
- Before output, you must first write # Analysis: explain task logic - which are query operations, which are state change operations - and how constraints are related.  

Strictly maintain the following Output Format:
# Analysis
[Explain operation requirements + classification logic + how constraints affect + ...]
# Operation List
## Information Query Class
- Operation: OperationName Description: ...
- Operation: OperationName Description: ...
- ...
## State Change Class
- Operation: OperationName Description: ...
- Operation: OperationName Description: ...
- ...
\end{lstlisting}
\vspace{-0.3cm}
\end{tcolorbox}
\caption{The prompt for planning tool operations of the environment.}
\label{fig:infer_operation_prompt}
\end{figure*}

\begin{figure*}
\begin{tcolorbox}[
colback=white!10!white,
colframe=black!75!black,
left=0.5pt, right=0.5pt,
boxsep=1pt,
title=SkelBuilder: Prompt for Converting State into Class Def,
]
\vspace{-0.3cm}
\begin{lstlisting}
You are an AI coding assistant. Your job is to translate an environment specification into a Python environment class definition. The class should simulate the stateful environment structure (without methods yet). You should analyze first and then generate code. You should follow the rules of Analysis and Code to generate the code.

Rules of Analysis
- Determine the environment class name. It should be EnvironmentSummary or an appropriate adaptation (e.g., LinuxFileSystem, EcommerceOrderSystem).  
- Extract attribute names (comma-separated) from each entity in state_space_definition.
- If needed, generate a corresponding TypedDict using the extracted attributes, with attribute name - key and attribute value type - inferred from the appropriate Python primitive type (e.g., id=str, name=str, category=str, price/size=float/int, quantity=int, status=str, timestamps=str/float).
- constraints_rules is left as a comment.

Rules of Code
- Generates each TypedDict definition if needed.
- Generates the environment class (with only __init__ and attributes), with attributes of type`Dict[ID, TypedDict].
- Add comments mapping each attribute back to the state space entity/attributes.
- Annotates the constraints in the code comments.
- Do not implement any business logic or methods yet.  
\end{lstlisting}
\vspace{-0.3cm}
\end{tcolorbox}
\caption{The prompt for programmatically converting states into the class definition.}
\label{fig:convert_state_to_class_prompt}
\end{figure*}

\begin{figure*}
\begin{tcolorbox}[
colback=white!10!white,
colframe=black!75!black,
left=0.5pt, right=0.5pt,
boxsep=1pt,
title=SkelBuilder: Prompt for Converting Tool to Class Method,
]
\vspace{-0.3cm}
\begin{lstlisting}
You are a code generation assistant.
Given an Agent's environment, including the environment's summary and introduction, the environment's state space definition, the environment's constraint rules, key base class definitions, and the list of operations supported by the environment.
Operations include two types: one is information querying of the environment, and the other is state modification of the environment.
Given one of the operations in the operation list (Target Operation),

You must:  
1. In # Analysis, reason about:  
   - What entities/attributes are involved.  
   - Parameters needed.  
   - Expected outputs (queries return structured results, state modifications return success messages).  
   - Error/edge cases (e.g., invalid input, permission denied).  
   - Does it involve environmental constraints or rules.  
2. In # Code, implement the Python method:  
   - Method name: `def <operation_name>(self, ...)`.  Note: Cannot be an independent function, but rather a method function within an already implemented environment class.
   - Add clear type hints.  
   - Add docstring describing inputs, outputs, constraints.  
   - Error handling: do not raise exceptions - return a dict like `{ "success": False, "error": "reason" }`.  
   - For information-query operations, if successful return `{ "success": True, "data": <result> }`.  
   - For state-modifying operations, if successful return `{ "success": True, "message": "operation description" }`. 

Your output format must be:
# Analysis
[Explain reasoning:inputs, outputs, related entities/attributes, constraints logic, success/failure cases]

# Code
```python
def <operation_name>(self, ...):
    """
    <docstring explaining inputs, outputs and constraints>
    """
    # Implementation
```
\end{lstlisting}
\vspace{-0.3cm}
\end{tcolorbox}
\caption{The prompt for programmatically converting operation to class-method.}
\label{fig:convert_op_to_func_prompt}
\end{figure*}

\begin{figure*}
\begin{tcolorbox}[
colback=white!10!white,
colframe=black!75!black,
left=0.5pt, right=0.5pt,
boxsep=1pt,
title=SkelBuilder: Prompt for Initializing Testing Agent,
]
\vspace{-0.3cm}
\begin{lstlisting}
You are an experienced testing engineer, performing comprehensive exploratory testing on all tool interfaces (methods) of a simulated environment class.  
Your goal is to verify the behavior of each method under different types of inputs, aiming to uncover potential errors, exceptions, and state inconsistencies.  
In each upcoming testing round, you will generate one tool invocation as a test case. After execution, you will receive the environment's return information, along with a result evaluation from backend engineers indicating the test case's result (pass, warning, fail).

[Environment Introduction]  
{env_introduction}

[Available Tool Interface List]  
{tool_info}

Testing Strategy:
- Positive case testing: Use valid parameters that comply with interface definitions (normal input).
- Negative case testing: Use invalid parameters (wrong types, non-existent IDs, out-of-range values, etc.) and special parameters (null/empty values, extreme values, special characters, etc.).
- Throughout testing, cover all available tool interfaces, and ensure a balance between the number of tests for each tool.
- It is not necessary to maintain a consistent task goal; you are free to explore various methods and scenarios.

Testing Rules:
- Invoke only one tool interface per round.
- Parameters must be in dictionary structure; parameter keys must be valid, but parameter values can be invalid or boundary inputs for testing.
- Do not call any methods outside of the provided tool interface list.
- Balance breadth (cover all available methods) and depth (multiple input scenarios for each method) during testing.
\end{lstlisting}
\vspace{-0.3cm}
\end{tcolorbox}
\caption{The prompt for initializing the testing agent during environments assessment.}
\label{fig:testing_agent_prompt}
\end{figure*}

\begin{figure*}
\begin{tcolorbox}[
colback=white!10!white,
colframe=black!75!black,
left=0.5pt, right=0.5pt,
boxsep=1pt,
title=SkelBuilder: Prompt for Initializing Checking Agent,
]
\vspace{-0.3cm}
\begin{lstlisting}
You are an experienced "Interactive Simulation Environment Testing Specialist", with extensive background in validating simulated systems and environments (such as game simulations, business system sandboxes, etc.).

Your task is to fully analyze whether a given method call in an environment class meets the expected behavior, based on the provided environment class structure, method source code, call parameters, the relevant internal state before and after the call, the differences between these states, and the method's returned observation.  

You should pay special attention to:  
- Whether the method causes the relevant internal state to change correctly before and after the call  
- Whether the code logic, conditional checks, state changes, and return value are consistent  
- Whether there are any unexpected exceptions or logical errors  

Strictly output your answer in the following format:
[Analysis]  
Your Step-by-step analysis.

[Result]  
Answer only one of the three words: 'Pass', 'Warning', or 'Fail', without any other words.  
- 'Pass' - The method fully meets expectations, implementation is correct, and no issues are found.  
- 'Warning' - The method works and meets functional expectations, but there are potential issues such as missing parameter validation, lack of boundary checks, absence of fallback mechanisms, or minor style/robustness problems.  
- 'Fail' - The method does not meet functional expectations, contains major logic errors, incorrect state changes, unhandled exceptions, or behaviors that violate environment rules.

[Error Reason]  
If the answer to 'Result' is 'Fail' or 'Warning', provide the reason you believe the error occurred and accordiing solutions.  
If the answer is 'Pass',  just output 'No error'.
\end{lstlisting}
\vspace{-0.3cm}
\end{tcolorbox}
\caption{The prompt for initializing the checking agent during environments assessment.}
\label{fig:checking_agent_prompt}
\end{figure*}

%% file: env_scen_synth_prompt.tex
\begin{figure*}
\begin{tcolorbox}[
colback=white!10!white,
colframe=black!75!black,
left=0.5pt, right=0.5pt,
boxsep=1pt,
title=ScenGenerator: Prompt for Generating Env Initial State,
]
\vspace{-0.3cm}
\begin{lstlisting}
You are an AI assistant.  
You will be given the complete definition of a Python class.  
This class represents an environment state in a specific domain and contains various attributes (such as dictionaries, lists, `TypedDict` objects, dataclasses, etc.) used to manage entities and their relationships within the system.
Based on the class definition, generate a JSON object that can serve directly as the class's initialization configuration (`config`), following these rules:

### 1. Structure and Type Matching  
- The JSON must strictly follow the attribute structure and data types required by the class.  
- Field names, nesting levels, and value types must match the class definition exactly.

### 2. Respect Constraints  
- Read the class methods and docstrings to identify constraints (e.g., valid status values, required fields, ID reference rules), ensuring all generated data complies.  
- All references (e.g., `reporter_id`, `location_id`, `disease_name`) must be cross-linked appropriately and valid.  
- Consider cross-entity relationships and constraints (e.g., a product must belong to an existing category).

### 3. Richness of Data  
- Each major dictionary-like attribute should contain multiple entities (recommended at least 3-5 entries) with differentiated content to avoid repetitive templates.  
- Cover the different states and value ranges supported by the class wherever possible.  
- Dates should be distributed over a reasonable time span to provide diversity.  
- Numerical fields (e.g., `case_count`) should vary in range to simulate realistic system data.

### 4. Realistic Simulation of Data  
- Name fields should use natural-language fictional content (e.g., `"Alice Chan"`, `"Central City District"`) rather than mechanical placeholders like `name1` or `user001`.  
- Description fields should be concise, natural, and logically consistent with the domain's context.  
- Date fields must be in ISO format (`YYYY-MM-DD`) or timestamps, with dates reasonably distributed in time.  
- ID fields may mix short codes (e.g., `LOC1`, `REP1`) and UUIDs, but all must be unique.  
- Data must be fictitious and must not contain any real-world personal or sensitive information.

### 5. Output Format  
- Output only the JSON, without any extra explanation.  
- The JSON must be a complete, ready-to-use initialization configuration that can be passed directly to the class constructor as the `config` parameter.
\end{lstlisting}
\vspace{-0.3cm}
\end{tcolorbox}
\caption{The prompt for generating environment’s initial state data.}
\label{fig:gen_init_state_prompt}
\end{figure*}

\begin{figure*}
\begin{tcolorbox}[
colback=white!10!white,
colframe=black!75!black,
left=0.5pt, right=0.5pt,
boxsep=1pt,
title=ScenGenerator: Prompt for Generating Env Task,
]
\vspace{-0.3cm}
\begin{lstlisting}
You are a task design expert, responsible for creating realistic, clear, and challenging tasks for a specific interactive environment. You do not need to consider how the task will be executed; another execution expert will be responsible for completing it.

# Environment Introduction:  {env_introduction}
# Environment State Definition:  {env_state_definition}
# Supported Operation Interfaces: {env_modify_operation}
# Environment Rules / Constraints:  {env_rule}
# Current Environment Initial State / Database:  {env_init_state}

# Task Design Requirements:
- Realism: The task must closely align with the current environment, reflect a plausible real-world scenario, and have a well-defined objective with business relevance.
- Feasibility: 
    i. The task must be based on and constrained by the current initial state of the environment. For example, you cannot delete a file that does not exist.
    ii. The task must be achievable through a combination of operation interfaces supported by the environment.
    iii. Avoid tasks that require more than just the user interface. For example, timestamps might be automatically generated rather than modified through the user interface.
- State Modification: The task must involve modifying the current environment state/database; it cannot be limited to information queries only.
- Challenge: The modification should have sufficient complexity (e.g., involving multiple objects, multiple attributes, or multi-condition combinations), and should not be achievable in a single simple step.
- Clarity: Use concise natural language to describe the task. The description must be easy to understand and unambiguous.
\end{lstlisting}
\vspace{-0.3cm}
\end{tcolorbox}
\caption{The prompt for generating a task under the specific environment and state.}
\label{fig:gen_task_prompt}
\end{figure*}

\begin{figure*}
\begin{tcolorbox}[
colback=white!10!white,
colframe=black!75!black,
left=0.5pt, right=0.5pt,
boxsep=1pt,
title=ScenGenerator: Prompt for Generating Verification Checklist,
]
\vspace{-0.3cm}
\begin{lstlisting}
You are a Quality Checklist Generation Assistant. I will provide you with a task description. Your job is to generate a simple and uniformly phrased checklist.
Requirements:
1. Each checklist item must be independent and not rely on the results of other items.  
2. Every checklist item must start with the exact phrase: "Has ..." followed by a clear description of the action or field to verify.  
3. Use precise fields and exact values; avoid vague wording.  
4. If the task requires checking multiple fields, split them into separate checklist items.  
5. List the items in logical order, ensuring each is self-contained.  
6. Output format:
   - Use Markdown list syntax (`- `) for each checklist item.  
   - Each item must start with "Has ..." and be verifiable with a single boolean expression.  
\end{lstlisting}
\vspace{-0.3cm}
\end{tcolorbox}
\vspace{-0.5cm}
\caption{The prompt for generating verification checklist for a task.}
\label{fig:gen_checklist_prompt}
\end{figure*}

\begin{figure*}
\begin{tcolorbox}[
colback=white!10!white,
colframe=black!75!black,
left=0.5pt, right=0.5pt,
boxsep=1pt,
title=ScenGenerator: Prompt for Converting Checkpoint to Verification Function,
]
\vspace{-0.3cm}
\begin{lstlisting}
You are a Python verification function generation assistant. You will be given:
- Environment introduction (env_introduction), describing the overall context and purpose of the system.
- Environment system configuration (initial_state), defining the structure, field names, and value formats of the system's data, and serving as the environment's initial database state before any modifications occur.
- After a series of modifications performed by an Agent to complete a task, the system reaches the final state (`final_state`).  
- A single check item, phrased as "Has ...", indicating the condition that must be verified against the final state.
Your task: 
Generate a Python function that validates whether the `final_state` satisfies the given check item.
Rules:
1. Always reference the data structure, field names, and value formats from `initial_state` when writing your verification logic. This ensures the code matches the actual system schema.
2. If the check item involves randomly generated or time-dependent fields (e.g., `user_id`, `create_time`, `update_time`, UUIDs), do not validate against a fixed concrete value.  
   Instead, check that the field exists in `final_state` and has the correct type/format (e.g., string).
3. If the check item describes a non-fixed target value (e.g., "add a remark"), only verify that the field exists and meets basic conditions (e.g., non-empty string, correct data type).
4. If the check item specifies an explicit target value, you must strictly match it (`==`).
5. Use `initial_state` as a reference only when necessary to determine changes - for example, "Has been added" means the entity didn't exist in `initial_state` but exists in `final_state`.
6. The function must implement only the given single check item and return `True` if passed, `False` if failed.
7. The function must not modify any input data and must perform no actions other than verification.
8. Ensure the function signature is `def check_func(final_state)`, and that it only returns `True` or `False`.
\end{lstlisting}
\vspace{-0.3cm}
\end{tcolorbox}
\caption{The prompt for programmatically converting each checkpoint to a verification function.}
\label{fig:gen_check_func_prompt}
\end{figure*}

%% file: llm_agent_prompt.tex
\begin{figure*}
\begin{tcolorbox}[
colback=white!10!white,
colframe=black!75!black,
left=0.5pt, right=0.5pt,
boxsep=1pt,
title=System Prompt for Non-Conversation Agent,
]
\vspace{-0.3cm}
\begin{lstlisting}
You are a helpful assistant. When given a specific task, your goal is to complete it in an interactive environment by making step-by-step use of available tools. 
- Before completing the task, at each step, select a tool from the tool list and fill in all required parameters, making sure that the values are valid. Avoid making parallel tool calls in one step.
- When you believe the task has been completed, respond only with 'Task Completed' to end the trajectory, without adding any other content or making any tool calls.
- It is recommended to first call query tools to gather sufficient information, then use modification tools to complete the task. Adjust actions promptly based on the feedback from the environment, i.e., the tool results.
\end{lstlisting}
\vspace{-0.3cm}
\end{tcolorbox}
\caption{The system prompt for prompting LLM agents under the \textit{Non-conversation} setting.}
\label{fig:agent_non_conv_prompt}
\end{figure*}

\begin{figure*}
\begin{tcolorbox}[
colback=white!10!white,
colframe=black!75!black,
left=0.5pt, right=0.5pt,
boxsep=1pt,
title=System Prompt for Conversation Agent,
]
\vspace{-0.3cm}
\begin{lstlisting}
You are a helpful assistant. Your goal is to fulfill the user's requests in an interactive environment by step-by-step use of available tools, while proactively communicating with the user when necessary, until the user ends the conversation.
At each step, you will receive either the user's task/reply or the environment's tool call result.
- If you lack essential information to complete the task or perform a tool call, and it cannot be obtained through the existing tool set, actively ask the user for specific details.  
- If you can proceed with the current information, select one tool from the tool set and provide complete, valid parameters. Avoid making parallel tool calls or calling a tool while interacting with the user in one step.
- It is recommended to first call query tools to gather sufficient information, then use modification tools to complete the task. Adjust actions promptly based on the feedback from the environment, i.e., the tool results.
- When you believe the task is completed, clearly inform the user of the result and ask whether there are any new tasks or follow-up requests. 
\end{lstlisting}
\vspace{-0.3cm}
\end{tcolorbox}
\caption{The system prompt for prompting LLM agents under the \textit{Conversation} setting.}
\label{fig:agent_conv_prompt}
\end{figure*}


\begin{figure*}
\begin{tcolorbox}[
colback=white!10!white,
colframe=black!75!black,
left=0.5pt, right=0.5pt,
boxsep=1pt,
title=System Prompt for LLM-Simulated User,
]
\vspace{-0.3cm}
\begin{lstlisting}
You are a real human user interacting with an Agent assistant.  
Your current task is to have the Agent accomplish the following goal:  
{task}  

Core Principles: 
- Do not directly or fully repeat the exact task instruction in your dialogue; instead, progress toward the goal gradually through multiple exchanges.  
- Deliver the task information in parts during the conversation so the Agent can slowly understand and move closer to the final objective.  
- When the task goal has been achieved, output a standalone message: `###STOP###` in your reply to end the dialogue. Do not include anything else.  

Rules:
1. If the task contains multiple sub-tasks, do not reveal all of them at once; provide relevant sub-tasks one by one as the Agent asks.  
2. If completing the task requires multiple pieces of information, do not disclose them all at once; provide partial information in response to the Agent's questions.  
3. All requests must remain strictly within the scope of the task-do not add extra requirements, intentions, or invent information that was not part of the original task.  
4. Always keep the conversation focused on progressing toward the task, ensuring every sub-task or goal is covered and none are skipped.  

Fidelity and Consistency Requirements: 
- Always remain faithful to the original task wording throughout the conversation. Pay special attention to preserving exact keywords, names, and proper nouns}@*)-do not rephrase or alter them.  
- If the Agent assistant presents you with multiple options, only choose those that match the intent and constraints of the original task. If none fit, politely refuse and restate your requirement.  
- Do notintroduce any new information that is not present in the original task description.  
- Do not repeat information you have already provided earlier in the conversation unless the Agent explicitly asks for clarification.  

Style Requirements: 
- Keep the dialogue natural and conversational, avoiding overly rigid or formal expressions.  

Output Format (must be strictly followed): 
# Thought:  
<Your thought process (this will NOT be sent to the Agent)> 
# Reply:  
<Your natural, conversational reply as the user, to be sent to the Agent>
\end{lstlisting}
\vspace{-0.3cm}
\end{tcolorbox}
\caption{The system prompt for prompting the LLM to act as a user under the \textit{Conversation} setting.}
\label{fig:sys_prompt_user}
\end{figure*}

%% file: tool_schema_example.tex
\begin{table}[t]
\scriptsize
\centering
\begin{tabularx}{\textwidth}{p{0.17\textwidth} >{\raggedright\arraybackslash}X p{0.18\textwidth}}
\toprule
\textbf{Tool Name} & \textbf{Description} & \textbf{Parameters} \\
\midrule
\makecell[tl]{\texttt{get\_user\_by\_phone\_} \\ \texttt{number}} & {\ttfamily Look up a user by their phone number. \newline Args: phone\_number (str): The phone number to search for.} & {\ttfamily phone\_number} \\
\midrule
\texttt{get\_user\_by\_id} & {\ttfamily Retrieve user information by user ID. \newline Args: user\_id (str): The unique ID of the user to be retrieved.} & {\ttfamily user\_id} \\
\midrule
\texttt{list\_all\_users} & {\ttfamily List all registered user accounts. \newline Args: None} & {\ttfamily None} \\
\midrule
\makecell[tl]{\texttt{get\_contact\_by\_phone\_} \\ \texttt{number}} & {\ttfamily Retrieve contact information using a contact's phone number. \newline Args: phone\_number (str): The phone number to look up in contacts.} & {\ttfamily phone\_number} \\
\midrule
\texttt{get\_contact\_by\_id} & {\ttfamily Look up and return contact info by contact ID. \newline Args: contact\_id (str): Unique identifier of the contact.} & {\ttfamily contact\_id} \\
\midrule
\texttt{list\_user\_contacts} & {\ttfamily Lists all contacts associated with a particular user. \newline Args: user\_id (str): The unique identifier for the user whose contacts are to be listed.} & {\ttfamily user\_id} \\
\midrule
\makecell[tl]{\texttt{validate\_phone\_} \\ \texttt{number}} & {\ttfamily Check if a phone number is valid for messaging. \newline Args: phone\_number (str): The phone number to validate.} & {\ttfamily phone\_number} \\
\midrule
\makecell[tl]{\texttt{get\_messages\_by\_} \\ \texttt{phone\_number}} & {\ttfamily Retrieve all messages (sent or received) associated with the specified phone number. \newline Args: phone\_number (str): The phone number for which to fetch messages.} & {\ttfamily phone\_number} \\
\midrule
\makecell[tl]{\texttt{get\_conversation\_by\_} \\ \texttt{user\_and\_contact}} & {\ttfamily Get the conversation history (messages) between a user and a given contact. \newline Args: user\_id (str): The user's id. contact\_id (str): The contact's id.} & {\ttfamily user\_id, contact\_id} \\
\midrule
\texttt{get\_message\_by\_id} & {\ttfamily Retrieve information for a specific message given its unique message ID. \newline Args: message\_id (str): The unique identifier of the message.} & {\ttfamily message\_id} \\
\midrule
\makecell[tl]{\texttt{get\_message\_delivery\_} \\ \texttt{status}} & {\ttfamily Check and return the current delivery status of a specific message. \newline Args: message\_id (str): The unique identifier of the message to query.} & {\ttfamily message\_id} \\
\midrule
\texttt{send\_message} & {\ttfamily Create and send a new message from a registered user to a valid phone number. \newline Args: sender\_id (str): Registered user's ID of the message sender. receiver\_phone\_number (str): The recipient's phone number. content (str): The textual content of the message.} & {\ttfamily sender\_id, receiver\_phone\_number, content} \\
\midrule
\makecell[tl]{\texttt{update\_message\_} \\ \texttt{delivery\_status}} & {\ttfamily Set or update the delivery status for a particular message. \newline Args: message\_id (str): The unique identifier of the message to update. new\_status (str): The new delivery status (e.g., 'sent', 'delivered', 'failed').} & {\ttfamily message\_id, new\_status} \\
\midrule
\texttt{create\_contact} & {\ttfamily Add a new contact to the specified user's contact list. \newline Args: contact\_id (str): The user id for whom the contact is being added. contact\_phone\_number (str): The contact's phone number. contact\_name (str): The contact's name.} & {\ttfamily contact\_id, contact\_phone\_number, contact\_name} \\
\midrule
\makecell[tl]{\texttt{add\_message\_to\_} \\ \texttt{conversation}} & {\ttfamily Insert a new message into the conversation history for the appropriate user-contact pair. \newline Args: conversation\_id (str): ID of the conversation history to add the message to. message\_id (str): ID of the message to insert (must exist and be valid).} & {\ttfamily conversation\_id, message\_id} \\
\midrule
\texttt{create\_conversation} & {\ttfamily Initiate a new conversation record between two users or a user and a contact, if such a conversation does not already exist. \newline Args: user\_id1 (str): The ID of the first user/participant. user\_id2 (str): The ID of the second user/participant.} & {\ttfamily user\_id1, user\_id2} \\
\midrule
\texttt{mark\_message\_as\_read} & {\ttfamily Mark a message as read by updating its read\_status field. \newline Args: message\_id (str): The unique identifier of the message to update.} & {\ttfamily message\_id} \\
\midrule
\texttt{delete\_message} & {\ttfamily Remove a message from the system. If message exists, it's deleted and any references in conversations are also removed. \newline Args: message\_id (str): Unique identifier of the message to remove.} & {\ttfamily message\_id} \\
\midrule
\texttt{archive\_conversation} & {\ttfamily Archive or hide a conversation for a specific user. \newline Args: user\_id (str): The user who wants to archive the conversation. conversation\_id (str): The conversation to archive.} & {\ttfamily user\_id, conversation\_id} \\
\midrule
\texttt{delete\_conversation} & {\ttfamily Remove a conversation from history if the action is allowed. \newline Args: conversation\_id (str): The unique ID of the conversation to remove. requestor\_id (str): The user requesting the deletion.} & {\ttfamily conversation\_id, requestor\_id} \\
\midrule
\end{tabularx}
\caption{An example of tool interfaces provided by synthesized environments to the LLM agent.}
\label{tab:tool_schema_example}
\end{table}

%% file: task_exmaple.tex
\begin{figure*}
\begin{tcolorbox}[
    enhanced jigsaw,
    colback=white,
    colframe=black!75!black,
    left=0.5pt, right=0.5pt,
    boxsep=1pt,
    title=Example of Environment's Initial State Configuration,
    breakable,
    fontupper=\scriptsize
]
\begin{lstlisting}[style=json]
{
    (*@\textcolor{jsongreen}{"users"}@*): {
        (*@\textcolor{jsongreen}{"USR1"}@*): {(*@\textcolor{jsongreen}{"\_id"}@*): "USR1", (*@\textcolor{jsongreen}{"phone\_number"}@*): "+15551234567", (*@\textcolor{jsongreen}{"name"}@*): "Alice Chan", (*@\textcolor{jsongreen}{"profile\_info"}@*): "admin"},
        (*@\textcolor{jsongreen}{"USR2"}@*): {(*@\textcolor{jsongreen}{"\_id"}@*): "USR2", (*@\textcolor{jsongreen}{"phone\_number"}@*): "+15559874566", (*@\textcolor{jsongreen}{"name"}@*): "Brandon Lee", (*@\textcolor{jsongreen}{"profile\_info"}@*): "basic user"},
        (*@\textcolor{jsongreen}{"USR3"}@*): {(*@\textcolor{jsongreen}{"\_id"}@*): "USR3", (*@\textcolor{jsongreen}{"phone\_number"}@*): "+15557678987", (*@\textcolor{jsongreen}{"name"}@*): "Cara Santorini", (*@\textcolor{jsongreen}{"profile\_info"}@*): "support"},
        (*@\textcolor{jsongreen}{"USR4"}@*): {(*@\textcolor{jsongreen}{"\_id"}@*): "USR4", (*@\textcolor{jsongreen}{"phone\_number"}@*): "+447531234762", (*@\textcolor{jsongreen}{"name"}@*): "David Newton", (*@\textcolor{jsongreen}{"profile\_info"}@*): "basic user"},
        (*@\textcolor{jsongreen}{"USR5"}@*): {(*@\textcolor{jsongreen}{"\_id"}@*): "USR5", (*@\textcolor{jsongreen}{"phone\_number"}@*): "+812345678901", (*@\textcolor{jsongreen}{"name"}@*): "Emi Takahashi", (*@\textcolor{jsongreen}{"profile\_info"}@*): "basic user"}
    },
    (*@\textcolor{jsongreen}{"contacts"}@*): {
        (*@\textcolor{jsongreen}{"contact\_1"}@*): {(*@\textcolor{jsongreen}{"\_id"}@*): "contact_1", (*@\textcolor{jsongreen}{"contact\_id"}@*): "USR1", (*@\textcolor{jsongreen}{"contact\_phone\_number"}@*): "+15559874566", (*@\textcolor{jsongreen}{"contact\_name"}@*): "Brandon Lee"},
        (*@\textcolor{jsongreen}{"contact\_2"}@*): {(*@\textcolor{jsongreen}{"\_id"}@*): "contact_2", (*@\textcolor{jsongreen}{"contact\_id"}@*): "USR1", (*@\textcolor{jsongreen}{"contact\_phone\_number"}@*): "+17165558888", (*@\textcolor{jsongreen}{"contact\_name"}@*): "Gabby Fields"},
        (*@\textcolor{jsongreen}{"contact\_3"}@*): {(*@\textcolor{jsongreen}{"\_id"}@*): "contact_3", (*@\textcolor{jsongreen}{"contact\_id"}@*): "USR2", (*@\textcolor{jsongreen}{"contact\_phone\_number"}@*): "+447531234762", (*@\textcolor{jsongreen}{"contact\_name"}@*): "David Newton"},
        (*@\textcolor{jsongreen}{"contact\_4"}@*): {(*@\textcolor{jsongreen}{"\_id"}@*): "contact_4", (*@\textcolor{jsongreen}{"contact\_id"}@*): "USR3", (*@\textcolor{jsongreen}{"contact\_phone\_number"}@*): "+812345678901", (*@\textcolor{jsongreen}{"contact\_name"}@*): "Emi Takahashi"},
        (*@\textcolor{jsongreen}{"contact\_5"}@*): {(*@\textcolor{jsongreen}{"\_id"}@*): "contact_5", (*@\textcolor{jsongreen}{"contact\_id"}@*): "USR3", (*@\textcolor{jsongreen}{"contact\_phone\_number"}@*): "+15557678987", (*@\textcolor{jsongreen}{"contact\_name"}@*): "Cara Santorini"}
    },
    (*@\textcolor{jsongreen}{"messages"}@*): {
        (*@\textcolor{jsongreen}{"MSG001"}@*): {(*@\textcolor{jsongreen}{"message\_id"}@*): "MSG001", (*@\textcolor{jsongreen}{"sender\_id"}@*): "USR1", (*@\textcolor{jsongreen}{"receiver\_id"}@*): "USR2", (*@\textcolor{jsongreen}{"receiver\_phone\_number"}@*): "+15559874566", (*@\textcolor{jsongreen}{"content"}@*): "Hey Brandon, can you send the report by Friday?", (*@\textcolor{jsongreen}{"timestamp"}@*): "2024-02-17T14:38:05Z", (*@\textcolor{jsongreen}{"delivery\_status"}@*): "delivered", (*@\textcolor{jsongreen}{"read\_status"}@*): "read"},
        (*@\textcolor{jsongreen}{"MSG002"}@*): {(*@\textcolor{jsongreen}{"message\_id"}@*): "MSG002", (*@\textcolor{jsongreen}{"sender\_id"}@*): "USR2", (*@\textcolor{jsongreen}{"receiver\_id"}@*): "USR1", (*@\textcolor{jsongreen}{"receiver\_phone\_number"}@*): "+15551234567", (*@\textcolor{jsongreen}{"content"}@*): "Hi Alice, yes I'll have it done tomorrow.", (*@\textcolor{jsongreen}{"timestamp"}@*): "2024-02-17T15:00:51Z", (*@\textcolor{jsongreen}{"delivery\_status"}@*): "delivered", (*@\textcolor{jsongreen}{"read\_status"}@*): "unread"},
        (*@\textcolor{jsongreen}{"MSG003"}@*): {(*@\textcolor{jsongreen}{"message\_id"}@*): "MSG003", (*@\textcolor{jsongreen}{"sender\_id"}@*): "USR3", (*@\textcolor{jsongreen}{"receiver\_id"}@*): "USR4", (*@\textcolor{jsongreen}{"receiver\_phone\_number"}@*): "+447531234762", (*@\textcolor{jsongreen}{"content"}@*): "David, tech support says the system update is finished.", (*@\textcolor{jsongreen}{"timestamp"}@*): "2024-03-05T11:21:12Z", (*@\textcolor{jsongreen}{"delivery\_status"}@*): "sent", (*@\textcolor{jsongreen}{"read\_status"}@*): "unread"},
        (*@\textcolor{jsongreen}{"MSG004"}@*): {(*@\textcolor{jsongreen}{"message\_id"}@*): "MSG004", (*@\textcolor{jsongreen}{"sender\_id"}@*): "USR4", (*@\textcolor{jsongreen}{"receiver\_id"}@*): "USR3", (*@\textcolor{jsongreen}{"receiver\_phone\_number"}@*): "+15557678987", (*@\textcolor{jsongreen}{"content"}@*): "Thank you Cara, I'll check and confirm later today.", (*@\textcolor{jsongreen}{"timestamp"}@*): "2024-03-05T11:31:58Z", (*@\textcolor{jsongreen}{"delivery\_status"}@*): "delivered", (*@\textcolor{jsongreen}{"read\_status"}@*): "read"},
        (*@\textcolor{jsongreen}{"MSG005"}@*): {(*@\textcolor{jsongreen}{"message\_id"}@*): "MSG005", (*@\textcolor{jsongreen}{"sender\_id"}@*): "USR2", (*@\textcolor{jsongreen}{"receiver\_id"}@*): "", (*@\textcolor{jsongreen}{"receiver\_phone\_number"}@*): "+17165558888", (*@\textcolor{jsongreen}{"content"}@*): "Gabby, please join the video call at 4pm.", (*@\textcolor{jsongreen}{"timestamp"}@*): "2024-04-02T08:45:19Z", (*@\textcolor{jsongreen}{"delivery\_status"}@*): "failed", (*@\textcolor{jsongreen}{"read\_status"}@*): "unread"},
        (*@\textcolor{jsongreen}{"MSG006"}@*): {(*@\textcolor{jsongreen}{"message\_id"}@*): "MSG006", (*@\textcolor{jsongreen}{"sender\_id"}@*): "USR5", (*@\textcolor{jsongreen}{"receiver\_id"}@*): "", (*@\textcolor{jsongreen}{"receiver\_phone\_number"}@*): "+15551234567", (*@\textcolor{jsongreen}{"content"}@*): "Alice, your settings have been updated securely.", (*@\textcolor{jsongreen}{"timestamp"}@*): "2024-06-01T10:02:46Z", (*@\textcolor{jsongreen}{"delivery\_status"}@*): "sent", (*@\textcolor{jsongreen}{"read\_status"}@*): "unread"},
        (*@\textcolor{jsongreen}{"MSG007"}@*): {(*@\textcolor{jsongreen}{"message\_id"}@*): "MSG007", (*@\textcolor{jsongreen}{"sender\_id"}@*): "USR1", (*@\textcolor{jsongreen}{"receiver\_id"}@*): "USR5", (*@\textcolor{jsongreen}{"receiver\_phone\_number"}@*): "+812345678901", (*@\textcolor{jsongreen}{"content"}@*): "Emi, the new version is available for testing.", (*@\textcolor{jsongreen}{"timestamp"}@*): "2024-05-15T07:17:22Z", (*@\textcolor{jsongreen}{"delivery\_status"}@*): "delivered", (*@\textcolor{jsongreen}{"read\_status"}@*): "read"}
    },
    (*@\textcolor{jsongreen}{"conversations"}@*): {
        (*@\textcolor{jsongreen}{"CONV001"}@*): {(*@\textcolor{jsongreen}{"conversation\_id"}@*): "CONV001", (*@\textcolor{jsongreen}{"user\_ids"}@*): ["USR1", "USR2"], (*@\textcolor{jsongreen}{"message\_id"}@*): "MSG002"},
        (*@\textcolor{jsongreen}{"CONV002"}@*): {(*@\textcolor{jsongreen}{"conversation\_id"}@*): "CONV002", (*@\textcolor{jsongreen}{"user\_ids"}@*): ["USR3", "USR4"], (*@\textcolor{jsongreen}{"message\_id"}@*): "MSG004"},
        (*@\textcolor{jsongreen}{"CONV003"}@*): {(*@\textcolor{jsongreen}{"conversation\_id"}@*): "CONV003", (*@\textcolor{jsongreen}{"user\_ids"}@*): ["USR2", "phone:+17165558888"], (*@\textcolor{jsongreen}{"message\_id"}@*): "MSG005"},
        (*@\textcolor{jsongreen}{"CONV004"}@*): {(*@\textcolor{jsongreen}{"conversation\_id"}@*): "CONV004", (*@\textcolor{jsongreen}{"user\_ids"}@*): ["USR1", "USR5"], (*@\textcolor{jsongreen}{"message\_id"}@*): "MSG007"},
        (*@\textcolor{jsongreen}{"CONV005"}@*): {(*@\textcolor{jsongreen}{"conversation\_id"}@*): "CONV005", (*@\textcolor{jsongreen}{"user\_ids"}@*): ["USR5", "phone:+15551234567"], (*@\textcolor{jsongreen}{"message\_id"}@*): "MSG006"}
    }
}
\end{lstlisting}
\end{tcolorbox}
\caption{An example of initial state data configuration for the environment.}
\label{fig:init_config_example}
\end{figure*}

\clearpage

\begin{figure*}
\begin{tcolorbox}[
colback=white!10!white,
colframe=black!75!black,
left=0.5pt, right=0.5pt,
boxsep=1pt,
    title=Task Example,
    breakable,
    fontupper=\footnotesize\ttfamily
]
Recover the failed outreach to Gabby Fields and report status: \\
1) As Brandon Lee (USR2), add Gabby Fields as a contact using her mobile number +17165558888. Validate the number before sending. \\
2) Re-send a corrected invite to Gabby from Brandon: “Hi Gabby, the meeting is now at 4:30pm ET. Please confirm.” Update the new message’s delivery status to “delivered.” \\
3) Link the new message to the existing Brandon–Gabby conversation and then archive that conversation. After successfully sending the new message, delete the old failed message Brandon previously sent to Gabby. \\
4) Send a short status update from Brandon to Alice Chan (USR1): “Update: re-sent Gabby’s invite and delivery confirmed.” Then mark Alice’s previously unread message from Brandon as “read.”
\end{tcolorbox}
\vspace{-0.5cm}
\caption{An example task under the above state configuration.}
\label{fig:task_example}
\end{figure*}

\begin{table*}
\scriptsize

\begin{tabular}{p{0.025\textwidth} p{0.14\textwidth} p{0.75\textwidth}}
\toprule
\textbf{No.} & \textbf{Checkpoint} & \textbf{Check Function} \\
\midrule
1 & Has Gabby Fields been added as a contact for user USR2 (Brandon Lee)? & \vspace{-3mm}
\begin{minipage}{\linewidth}
\begin{lstlisting}[style=python, basicstyle=\ttfamily\scriptsize, breaklines=true, frame=none, aboveskip=3pt, belowskip=3pt]
def check_func(final_state):
    for contact in final_state.get("contacts", {}).values():
        if (contact.get("contact_id") == "USR2" and
            contact.get("contact_phone_number") == "+17165558888" and
            contact.get("contact_name") == "Gabby Fields"):
            return True
    return False
\end{lstlisting}
\end{minipage} \\
\midrule
2 & Has the mobile number of Gabby Fields been set to "+17165558888"? & \vspace{-3mm}
\begin{minipage}{\linewidth}
\begin{lstlisting}[style=python, basicstyle=\ttfamily\scriptsize, breaklines=true, frame=none, aboveskip=3pt, belowskip=3pt]
def check_func(final_state):
    for contact in final_state.get("contacts", {}).values():
        if contact.get("contact_name") == "Gabby Fields" and contact.get("contact_phone_number") == "+17165558888":
            return True
    return False
\end{lstlisting}
\end{minipage} \\
\midrule
3 & Has the new corrected message from USR2 to Gabby Fields been created with text "Hi Gabby, the meeting is now at 4:30pm ET. Please confirm."? & \vspace{-3mm}
\begin{minipage}{\linewidth}
\begin{lstlisting}[style=python, basicstyle=\ttfamily\scriptsize, breaklines=true, frame=none, aboveskip=3pt, belowskip=3pt]
def check_func(final_state):
    target_text = "Hi Gabby, the meeting is now at 4:30pm ET. Please confirm."
    for msg in final_state.get("messages", {}).values():
        if (msg.get("sender_id") == "USR2" and msg.get("receiver_id") == ""
            and msg.get("receiver_phone_number") == "+17165558888"
            and msg.get("content") == target_text):
            return True
    return False
\end{lstlisting}
\end{minipage} \\
\midrule
4 & Has the delivery\_status of the new corrected message been set to "delivered"? & \vspace{-3mm}
\begin{minipage}{\linewidth}
\begin{lstlisting}[style=python, basicstyle=\ttfamily\scriptsize, breaklines=true, frame=none, aboveskip=3pt, belowskip=3pt]
def check_func(final_state):
    target_content = "Hi Gabby, the meeting is now at 4:30pm ET. Please confirm."
    for msg in final_state.get("messages", {}).values():
        if (msg.get("sender_id") == "USR2"
            and msg.get("receiver_phone_number") == "+17165558888"
            and msg.get("content") == target_content
            and msg.get("delivery_status") == "delivered"):
            return True
    return False
\end{lstlisting}
\end{minipage} \\
\midrule
5 & Has the new corrected message been linked to the existing Brandon–Gabby conversation? & \vspace{-3mm}
\begin{minipage}{\linewidth}
\begin{lstlisting}[style=python, basicstyle=\ttfamily\scriptsize, breaklines=true, frame=none, aboveskip=3pt, belowskip=3pt]
def check_func(final_state):
    conv = final_state.get("conversations", {}).get("CONV003")
    if not conv:
        return False
    return conv.get("message_id") != "MSG005"
\end{lstlisting}
\end{minipage} \\
\midrule
6 & Has the Brandon–Gabby conversation been archived? & \vspace{-3mm}
\begin{minipage}{\linewidth}
\begin{lstlisting}[style=python, basicstyle=\ttfamily\scriptsize, breaklines=true, frame=none, aboveskip=3pt, belowskip=3pt]
def check_func(final_state):
    target_users = {"USR2", "phone:+17165558888"}
    for conv in final_state.get("conversations", {}).values():
        if set(conv.get("user_ids", [])) == target_users:
            return conv.get("archived") is True
    return False
\end{lstlisting}
\end{minipage} \\
\midrule
7 & Has the old failed message previously sent by USR2 to Gabby Fields been deleted? & \vspace{-3mm}
\begin{minipage}{\linewidth}
\begin{lstlisting}[style=python, basicstyle=\ttfamily\scriptsize, breaklines=true, frame=none, aboveskip=3pt, belowskip=3pt]
def check_func(final_state):
    return "MSG005" not in final_state.get("messages", {})
\end{lstlisting}
\end{minipage} \\
\midrule
8 & Has the status-update message from USR2 to Alice Chan (USR1) been created with text "Update: re-sent Gabby's invite and delivery confirmed."? & \vspace{-3mm}
\begin{minipage}{\linewidth}
\begin{lstlisting}[style=python, basicstyle=\ttfamily\scriptsize, breaklines=true, frame=none, aboveskip=3pt, belowskip=3pt]
def check_func(final_state):
    target_text = "Update: re-sent Gabby's invite and delivery confirmed."
    for msg in final_state.get("messages", {}).values():
        if msg.get("sender_id") == "USR2" and msg.get("receiver_id") == "USR1" and msg.get("content") == target_text:
            return True
    return False
\end{lstlisting}
\end{minipage} \\
\midrule
9 & Has Alice Chan's previously unread message from USR2 been marked as "read"? & \vspace{-3mm}
\begin{minipage}{\linewidth}
\begin{lstlisting}[style=python, basicstyle=\ttfamily\scriptsize, breaklines=true, frame=none, aboveskip=3pt, belowskip=3pt]
def check_func(final_state):
    msg = final_state["messages"].get("MSG002")
    return msg.get("read_status") == "read" if msg else False
\end{lstlisting}
\end{minipage} \\
\bottomrule
\end{tabular}
\vspace{-0.3cm}
\caption{The checklist and state-check functions corresponding to the task.}
\label{tab:check_func_example}
\end{table*}

%% file: traj_example.tex
\clearpage
\onecolumn

{\scriptsize
\begin{longtable}{p{0.032\textwidth} >{\raggedright\arraybackslash}p{0.9\textwidth}}
\toprule
\textbf{Turn} & \textbf{Interaction} \\
\midrule
\endhead
1 & \textbf{Observation}: {\ttfamily \textbf{[Task]} Recover the failed outreach to Gabby Fields and report status:

1) As Brandon Lee (USR2), add Gabby Fields as a contact using her mobile number +17165558888. Validate the number before sending.

2) Re-send a corrected invite to Gabby from Brandon: “Hi Gabby, the meeting is now at 4:30pm ET. Please confirm.” Update the new message’s delivery status to “delivered.”

3) Link the new message to the existing Brandon–Gabby conversation and then archive that conversation. After successfully sending the new message, delete the old failed message Brandon previously sent to Gabby.

4) Send a short status update from Brandon to Alice Chan (USR1): “Update: re-sent Gabby’s invite and delivery confirmed.” Then mark Alice’s previously unread message from Brandon as “read.”} \newline \textbf{Action}: {\ttfamily Function(name="validate\_phone\_number", arguments=\{"phone\_number": "+17165558888"\})} \\
\midrule
2 & \textbf{Observation}: {\ttfamily \textbf{[Tool Result]} \{'success': True, 'data': \{'valid': True, 'reason': 'Phone number is valid for messaging'\}\}} \newline \textbf{Action}: {\ttfamily Function(name="create\_contact", arguments=\{"contact\_id": "USR2", "contact\_phone\_number": "+17165558888", "contact\_name": "Gabby Fields"\})} \\
\midrule
3 & \textbf{Observation}: {\ttfamily \textbf{[Tool Result]} \{'success': True, 'message': 'Contact created'\}} \newline \textbf{Action}: {\ttfamily Function(name="get\_contact\_by\_phone\_number", arguments=\{"phone\_number": "+17165558888"\})} \\
\midrule
4 & \textbf{Observation}: {\ttfamily \textbf{[Tool Result]} \{'success': True, 'data': \{'\_id': 'contact\_2', 'contact\_id': 'USR1', 'contact\_phone\_number': '+17165558888', 'contact\_name': 'Gabby Fields'\}\}} \newline \textbf{Action}: {\ttfamily Function(name="get\_conversation\_by\_user\_and\_contact", arguments=\{"user\_id": "USR2", "contact\_id": "contact\_2"\})} \\
\midrule
5 & \textbf{Observation}: {\ttfamily \textbf{[Tool Result]} \{'success': False, 'error': 'No conversation found between user and contact'\}} \newline \textbf{Action}: {\ttfamily Function(name="get\_user\_by\_phone\_number", arguments=\{"phone\_number": "+17165558888"\})} \\
\midrule
6 & \textbf{Observation}: {\ttfamily \textbf{[Tool Result]} \{'success': False, 'error': 'User with this phone number not found'\}} \newline \textbf{Action}: {\ttfamily Function(name="create\_conversation", arguments=\{"user\_id1": "USR2", "user\_id2": "contact\_2"\})} \\
\midrule
7 & \textbf{Observation}: {\ttfamily \textbf{[Tool Result]} \{'success': True, 'message': 'Conversation created', 'conversation': \{'conversation\_id': '50aecb26-277a-478f-b9fe-cf5d89330781', 'user\_ids': ['USR2', 'contact\_2'], 'message\_id': ''\}\}} \newline \textbf{Action}: {\ttfamily Function(name="send\_message", arguments=\{"sender\_id": "USR2", "receiver\_phone\_number": "+17165558888", "content": "Hi Gabby, the meeting is now at 4:30pm ET. Please confirm."\})} \\
\midrule
8 & \textbf{Observation}: {\ttfamily \textbf{[Tool Result]} \{'success': True, 'message': 'Message sent successfully.'\}} \newline \textbf{Action}: {\ttfamily Function(name="get\_messages\_by\_phone\_number", arguments=\{"phone\_number": "+17165558888"\})} \\
\midrule
9 & \textbf{Observation}: {\ttfamily \textbf{[Tool Result]} \{'success': True, 'data': [\{'message\_id': 'MSG005', 'sender\_id': 'USR2', 'receiver\_id': '', 'receiver\_phone\_number': '+17165558888', 'content': 'Gabby, please join the video call at 4pm.', 'timestamp': '2024-04-02T08:45:19Z', 'delivery\_status': 'failed', 'read\_status': 'unread'\}, \{'message\_id': '5b5f4580-f7db-4550-b6bd-0d7ef221741f', 'sender\_id': 'USR2', 'receiver\_id': '', 'receiver\_phone\_number': '+17165558888', 'content': 'Hi Gabby, the meeting is now at 4:30pm ET. Please confirm.', 'timestamp': '1763026165.0235567', 'delivery\_status': 'sent', 'read\_status': 'unread'\}]\}} \newline \textbf{Action}: {\ttfamily Function(name="update\_message\_delivery\_status", arguments=\{"message\_id": "5b5f4580-f7db-4550-b6bd-0d7ef221741f", "new\_status": "delivered"\})} \\
\midrule
10 & \textbf{Observation}: {\ttfamily \textbf{[Tool Result]} \{'success': True, 'message': 'Delivery status updated for message 5b5f4580-f7db-4550-b6bd-0d7ef221741f'\}} \newline \textbf{Action}: {\ttfamily Function(name="add\_message\_to\_conversation", arguments=\{"conversation\_id": "50aecb26-277a-478f-b9fe-cf5d89330781", "message\_id": "5b5f4580-f7db-4550-b6bd-0d7ef221741f"\})} \\
\midrule
11 & \textbf{Observation}: {\ttfamily \textbf{[Tool Result]} \{'success': True, 'message': 'Message added to conversation history.'\}} \newline \textbf{Action}: {\ttfamily Function(name="archive\_conversation", arguments=\{"user\_id": "USR2", "conversation\_id": "50aecb26-277a-478f-b9fe-cf5d89330781"\})} \\
\midrule
12 & \textbf{Observation}: {\ttfamily \textbf{[Tool Result]} \{'success': True, 'message': 'Conversation archived for user.'\}} \newline \textbf{Action}: {\ttfamily Function(name="delete\_message", arguments=\{"message\_id": "MSG005"\})} \\
\midrule
13 & \textbf{Observation}: {\ttfamily \textbf{[Tool Result]} \{'success': True, 'message': 'Message MSG005 has been deleted.'\}} \newline \textbf{Action}: {\ttfamily Function(name="get\_user\_by\_id", arguments=\{"user\_id": "USR1"\})} \\
\midrule
14 & \textbf{Observation}: {\ttfamily \textbf{[Tool Result]} \{'success': True, 'data': \{'\_id': 'USR1', 'phone\_number': '+15551234567', 'name': 'Alice Chan', 'profile\_info': 'admin'\}\}} \newline \textbf{Action}: {\ttfamily Function(name="send\_message", arguments=\{"sender\_id": "USR2", "receiver\_phone\_number": "+15551234567", "content": "Update: re-sent Gabby’s invite and delivery confirmed."\})} \\
\midrule
15 & \textbf{Observation}: {\ttfamily \textbf{[Tool Result]} \{'success': True, 'message': 'Message sent successfully.'\}} \newline \textbf{Action}: {\ttfamily Function(name="get\_messages\_by\_phone\_number", arguments=\{"phone\_number": "+15551234567"\})} \\
\midrule
16 & \textbf{Observation}: {\ttfamily \textbf{[Tool Result]} \{'success': True, 'data': [\{'message\_id': 'MSG001', 'sender\_id': 'USR1', 'receiver\_id': 'USR2', 'receiver\_phone\_number': '+15559874566', 'content': 'Hey Brandon, can you send the report by Friday?', 'timestamp': '2024-02-17T14:38:05Z', 'delivery\_status': 'delivered', 'read\_status': 'read'\}, \{'message\_id': 'MSG002', 'sender\_id': 'USR2', 'receiver\_id': 'USR1', 'receiver\_phone\_number': '+15551234567', 'content': "Hi Alice, yes I'll have it done tomorrow.", 'timestamp': '2024-02-17T15:00:51Z', 'delivery\_status': 'delivered', 'read\_status': 'unread'\}, \{'message\_id': 'MSG006', 'sender\_id': 'USR5', 'receiver\_id': '', 'receiver\_phone\_number': '+15551234567', 'content': 'Alice, your settings have been updated securely.', 'timestamp': '2024-06-01T10:02:46Z', 'delivery\_status': 'sent', 'read\_status': 'unread'\}, \{'message\_id': 'MSG007', 'sender\_id': 'USR1', 'receiver\_id': 'USR5', 'receiver\_phone\_number': '+812345678901', 'content': 'Emi, the new version is available for testing.', 'timestamp': '2024-05-15T07:17:22Z', 'delivery\_status': 'delivered', 'read\_status': 'read'\}, \{'message\_id': '08e0a7b1-5039-433d-9dfd-7f7e4359d548', 'sender\_id': 'USR2', 'receiver\_id': 'USR1', 'receiver\_phone\_number': '+15551234567', 'content': 'Update: re-sent Gabby’s invite and delivery confirmed.', 'timestamp': '1763026220.1726387', 'delivery\_status': 'sent', 'read\_status': 'unread'\}]\}} \newline \textbf{Action}: {\ttfamily Function(name="mark\_message\_as\_read", arguments=\{"message\_id": "MSG002"\})} \\
\midrule
17 & \textbf{Observation}: {\ttfamily \textbf{[Tool Result]} \{'success': True, 'message': 'Message marked as read.'\}} \newline \textbf{Action}: {\ttfamily Task Completed} \\
\bottomrule
\caption{A trajectory example of Non-Conversation setting (The reasoning process <think>...</think> is ignored).}
\label{tab:non_conv_traj_example}
\end{longtable}
}
{\scriptsize
\begin{longtable}{p{0.03\textwidth} >{\raggedright\arraybackslash}p{0.9\textwidth}}
\toprule
\textbf{Turn} & \textbf{Interaction} \\
\midrule
\endhead
1 & \textbf{Observation}: {\ttfamily \textbf{[User]} Hey! I need some help with an outreach issue involving Gabby Fields. Can you assist me with getting her contact added?} \newline \textbf{Action}: {\ttfamily To add Gabby Fields as a contact, I'll need her phone number. Could you please provide Gabby's phone number so I can validate it and add her to your contact list?} \\
\midrule
2 & \textbf{Observation}: {\ttfamily \textbf{[User]} Sure, Gabby's mobile number is +17165558888. Let me know once you’ve validated it and added her to my contacts!} \newline \textbf{Action}: {\ttfamily Function(name="validate\_phone\_number", arguments=\{"phone\_number": "+17165558888"\})} \\
\midrule
3 & \textbf{Observation}: {\ttfamily \textbf{[Tool Result]} \{'success': True, 'data': \{'valid': True, 'reason': 'Phone number is valid for messaging'\}\}} \newline \textbf{Action}: {\ttfamily The phone number **+17165558888** has been validated as valid for messaging. To proceed with adding Gabby Fields to your contacts, I'll need your **user ID** (the unique identifier associated with your account). Could you please share your user ID so I can complete this request?} \\
\midrule
4 & \textbf{Observation}: {\ttfamily \textbf{[User]} You can use my user ID: Brandon Lee (USR2) for this. Please go ahead and add Gabby Fields using that info.} \newline \textbf{Action}: {\ttfamily Function(name="create\_contact", arguments=\{"contact\_id": "USR2", "contact\_phone\_number": "+17165558888", "contact\_name": "Gabby Fields"\})} \\
\midrule
5 & \textbf{Observation}: {\ttfamily \textbf{[Tool Result]} \{'success': True, 'message': 'Contact created'\}} \newline \textbf{Action}: {\ttfamily Gabby Fields (phone: +17165558888) has been successfully added to your contacts as **Gabby Fields**. Your contact list now includes this entry under your account (USR2). Let me know if you'd like to send a message to Gabby or need further assistance!} \\
\midrule
6 & \textbf{Observation}: {\ttfamily \textbf{[User]} Great, thanks! Next, I want to send Gabby a new invite from Brandon. The message should be: “Hi Gabby, the meeting is now at 4:30pm ET. Please confirm.” Please send it and update the delivery status to “delivered.”} \newline \textbf{Action}: {\ttfamily Function(name="send\_message", arguments=\{"sender\_id": "USR2", "receiver\_phone\_number": "+17165558888", "content": "Hi Gabby, the meeting is now at 4:30pm ET. Please confirm."\})} \\
\midrule
7 & \textbf{Observation}: {\ttfamily \textbf{[Tool Result]} \{'success': True, 'message': 'Message sent successfully.'\}} \newline \textbf{Action}: {\ttfamily Function(name="update\_message\_delivery\_status", arguments=\{"message\_id": "MSG123", "new\_status": "delivered"\})} \\
\midrule
8 & \textbf{Observation}: {\ttfamily \textbf{[Tool Result]} \{'success': False, 'error': 'Message not found'\}} \newline \textbf{Action}: {\ttfamily Function(name="get\_messages\_by\_phone\_number", arguments=\{"phone\_number": "+17165558888"\})} \\
\midrule
9 & \textbf{Observation}: {\ttfamily \textbf{[Tool Result]} \{'success': True, 'data': [\{'message\_id': 'MSG005', 'sender\_id': 'USR2', 'receiver\_id': '', 'receiver\_phone\_number': '+17165558888', 'content': 'Gabby, please join the video call at 4pm.', 'timestamp': '2024-04-02T08:45:19Z', 'delivery\_status': 'failed', 'read\_status': 'unread'\}, \{'message\_id': 'c02f3c58-a015-40ee-b034-390b649c472e', 'sender\_id': 'USR2', 'receiver\_id': '', 'receiver\_phone\_number': '+17165558888', 'content': 'Hi Gabby, the meeting is now at 4:30pm ET. Please confirm.', 'timestamp': '1763054273.124194', 'delivery\_status': 'sent', 'read\_status': 'unread'\}]\}} \newline \textbf{Action}: {\ttfamily Function(name="update\_message\_delivery\_status", arguments=\{"message\_id": "c02f3c58-a015-40ee-b034-390b649c472e", "new\_status": "delivered"\})} \\
\midrule
10 & \textbf{Observation}: {\ttfamily \textbf{[Tool Result]} \{'success': True, 'message': 'Delivery status updated for message c02f3c58-a015-40ee-b034-390b649c472e'\}} \newline \textbf{Action}: {\ttfamily The message to Gabby Fields has been successfully sent and the delivery status updated to **"delivered"**. Here's the confirmation:

**Message Content:**  
"Hi Gabby, the meeting is now at 4:30pm ET. Please confirm."

**Delivery Status:** Delivered  
**Message ID:** c02f3c58-a015-40ee-b034-390b649c472e

Let me know if you'd like to send another message or need further assistance!} \\
\midrule
11 & \textbf{Observation}: {\ttfamily \textbf{[User]} Perfect! Can you please link this new message to the existing Brandon–Gabby conversation? Once that’s done, go ahead and archive the whole conversation for me.} \newline \textbf{Action}: {\ttfamily Function(name="create\_conversation", arguments=\{"user\_id1": "USR2", "user\_id2": "+17165558888"\})} \\
\midrule
12 & \textbf{Observation}: {\ttfamily \textbf{[Tool Result]} \{'success': False, 'error': 'Participant 2 (+17165558888) does not exist.'\}} \newline \textbf{Action}: {\ttfamily Function(name="get\_contact\_by\_phone\_number", arguments=\{"phone\_number": "+17165558888"\})} \\
\midrule
13 & \textbf{Observation}: {\ttfamily \textbf{[Tool Result]} \{'success': True, 'data': \{'\_id': 'contact\_2', 'contact\_id': 'USR1', 'contact\_phone\_number': '+17165558888', 'contact\_name': 'Gabby Fields'\}\}} \newline \textbf{Action}: {\ttfamily Function(name="create\_conversation", arguments=\{"user\_id1": "USR2", "user\_id2": "contact\_2"\})} \\
\midrule
14 & \textbf{Observation}: {\ttfamily \textbf{[Tool Result]} \{'success': True, 'message': 'Conversation created', 'conversation': \{'conversation\_id': 'c3d610bf-1100-405f-8897-8df5cea96b1d', 'user\_ids': ['USR2', 'contact\_2'], 'message\_id': ''\}\}} \newline \textbf{Action}: {\ttfamily Function(name="add\_message\_to\_conversation", arguments=\{"conversation\_id": "c3d610bf-1100-405f-8897-8df5cea96b1d", "message\_id": "c02f3c58-a015-40ee-b034-390b649c472e"\})} \\
\midrule
15 & \textbf{Observation}: {\ttfamily \textbf{[Tool Result]} \{'success': True, 'message': 'Message added to conversation history.'\}} \newline \textbf{Action}: {\ttfamily Function(name="archive\_conversation", arguments=\{"user\_id": "USR2", "conversation\_id": "c3d610bf-1100-405f-8897-8df5cea96b1d"\})} \\
\midrule
16 & \textbf{Observation}: {\ttfamily \textbf{[Tool Result]} \{'success': True, 'message': 'Conversation archived for user.'\}} \newline \textbf{Action}: {\ttfamily The message has been successfully linked to the Brandon–Gabby conversation (ID: `c3d610bf-1100-405f-8897-8df5cea96b1d`), and the entire conversation has been archived for your account (USR2). 

**Summary:**  
Message "Hi Gabby, the meeting is now at 4:30pm ET. Please confirm." linked to conversation  
Conversation archived for future reference  

Let me know if you'd like to unarchive or take any other action!} \\
\midrule
17 & \textbf{Observation}: {\ttfamily \textbf{[User]} Thanks! Now, please delete the old failed message Brandon sent to Gabby from that conversation. Let me know once it’s removed.} \newline \textbf{Action}: {\ttfamily Function(name="delete\_message", arguments=\{"message\_id": "MSG005"\})} \\
\midrule
18 & \textbf{Observation}: {\ttfamily \textbf{[Tool Result]} \{'success': True, 'message': 'Message MSG005 has been deleted.'\}} \newline \textbf{Action}: {\ttfamily The old failed message (ID: `MSG005`) has been successfully deleted from the conversation. The conversation now only includes the confirmed message:  
**"Hi Gabby, the meeting is now at 4:30pm ET. Please confirm."**  

Let me know if you need further cleanup or assistance!} \\
\midrule
19 & \textbf{Observation}: {\ttfamily \textbf{[User]} Great, thank you. Lastly, send this short status update from Brandon to Alice Chan: “Update: re-sent Gabby’s invite and delivery confirmed.” And please mark Alice’s previously unread message from Brandon as “read.” Let me know when that’s done.} \newline \textbf{Action}: {\ttfamily To send the status update to Alice Chan and mark her unread message as read, I'll need **Alice Chan's phone number** to proceed. Could you please provide her phone number so I can validate it, send the message, and update the read status?} \\
\midrule
20 & \textbf{Observation}: {\ttfamily \textbf{[User]} I don't have Alice's phone number available, but her user ID is Alice Chan (USR1). Can you go ahead and send the update to her using that, and mark her unread message from Brandon as “read”?} \newline \textbf{Action}: {\ttfamily Function(name="get\_user\_by\_id", arguments=\{"user\_id": "USR1"\})} \\
\midrule
21 & \textbf{Observation}: {\ttfamily \textbf{[Tool Result]} \{'success': True, 'data': \{'\_id': 'USR1', 'phone\_number': '+15551234567', 'name': 'Alice Chan', 'profile\_info': 'admin'\}\}} \newline \textbf{Action}: {\ttfamily Function(name="send\_message", arguments=\{"sender\_id": "USR2", "receiver\_phone\_number": "+15551234567", "content": "Update: re-sent Gabby’s invite and delivery confirmed."\})} \\
\midrule
22 & \textbf{Observation}: {\ttfamily \textbf{[Tool Result]} \{'success': True, 'message': 'Message sent successfully.'\}} \newline \textbf{Action}: {\ttfamily Function(name="get\_messages\_by\_phone\_number", arguments=\{"phone\_number": "+15551234567"\})} \\
\midrule
23 & \textbf{Observation}: {\ttfamily \textbf{[Tool Result]} \{'success': True, 'data': [\{'message\_id': 'MSG001', 'sender\_id': 'USR1', 'receiver\_id': 'USR2', 'receiver\_phone\_number': '+15559874566', 'content': 'Hey Brandon, can you send the report by Friday?', 'timestamp': '2024-02-17T14:38:05Z', 'delivery\_status': 'delivered', 'read\_status': 'read'\}, \{'message\_id': 'MSG002', 'sender\_id': 'USR2', 'receiver\_id': 'USR1', 'receiver\_phone\_number': '+15551234567', 'content': "Hi Alice, yes I'll have it done tomorrow.", 'timestamp': '2024-02-17T15:00:51Z', 'delivery\_status': 'delivered', 'read\_status': 'unread'\}, \{'message\_id': 'MSG006', 'sender\_id': 'USR5', 'receiver\_id': '', 'receiver\_phone\_number': '+15551234567', 'content': 'Alice, your settings have been updated securely.', 'timestamp': '2024-06-01T10:02:46Z', 'delivery\_status': 'sent', 'read\_status': 'unread'\}, \{'message\_id': 'MSG007', 'sender\_id': 'USR1', 'receiver\_id': 'USR5', 'receiver\_phone\_number': '+812345678901', 'content': 'Emi, the new version is available for testing.', 'timestamp': '2024-05-15T07:17:22Z', 'delivery\_status': 'delivered', 'read\_status': 'read'\}, \{'message\_id': 'b2e86264-96cb-40c0-b7dd-39091b185d15', 'sender\_id': 'USR2', 'receiver\_id': 'USR1', 'receiver\_phone\_number': '+15551234567', 'content': 'Update: re-sent Gabby’s invite and delivery confirmed.', 'timestamp': '1763054573.840822', 'delivery\_status': 'sent', 'read\_status': 'unread'\}]\}} \newline \textbf{Action}: {\ttfamily Function(name="mark\_message\_as\_read", arguments=\{"message\_id": "MSG002"\})} \\
\midrule
24 & \textbf{Observation}: {\ttfamily \textbf{[Tool Result]} \{'success': True, 'message': 'Message marked as read.'\}} \newline \textbf{Action}: {\ttfamily The message to Alice Chan has been successfully sent:  
**"Update: re-sent Gabby’s invite and delivery confirmed."**  

Additionally, Alice’s previously unread message from Brandon (ID: `MSG002`) has been marked as **read**.  

All actions completed! Let me know if you need further assistance.} \\
\bottomrule
\caption{A trajectory example of Conversation setting (The reasoning process <think>...</think> is ignored).}
\label{tab:conv_traj_example}
\end{longtable}
}
\twocolumn
\clearpage